\RequirePackage{fix-cm}
\documentclass{svjour3-custom}                     

\smartqed  
\usepackage{graphicx}
\usepackage{epstopdf}

\usepackage{booktabs}

\usepackage{verbatim}
\usepackage{algorithm}
\usepackage{algpseudocode}
\usepackage{xspace}
\usepackage{amsmath}
\usepackage{caption}
\usepackage{subcaption}

\usepackage{rotating}
\usepackage{verbatimbox}

\journalname{\emph{Genetic Programming and Evolvable Machines},}
\begin{document}

\makeatletter
\DeclareRobustCommand\onedot{\futurelet\@let@token\@onedot}
\def\@onedot{\ifx\@let@token.\else.\null\fi\xspace}
\def\eg{\emph{e.g}\onedot} \def\Eg{\emph{E.g}\onedot}
\def\ie{\emph{i.e}\onedot} \def\Ie{\emph{I.e}\onedot}
\def\etal{\emph{et al}\onedot}
\makeatother

\title{An Automatic Solver for Very Large Jigsaw Puzzles Using Genetic Algorithms\footnote{A preliminary version of this paper appeared in  \emph{Proceedings of the IEEE Computer Vision and Pattern Recognition Conference} \cite{Sholomon_2013_CVPR}}}



\author{Dror Sholomon \and Eli (Omid) David \and Nathan S. Netanyahu }

\authorrunning{D. Sholomon, E.O. David, N.S. Netanyahu}

\institute{
D. Sholomon \at
Department of Computer Science, Bar-Ilan University, 52900 Ramat-Gan, Israel \\
\email{dror.sholomon@gmail.com}
\and
E.O. David \at
Department of Computer Science, Bar-Ilan University, 52900 Ramat-Gan, Israel \\
\email{mail@elidavid.com}, Website: www.elidavid.com
\and
N.S. Netanyahu \at
Department of Computer Science, Bar-Ilan University, 52900 Ramat-Gan, Israel and\\
Center for Automation Research, University of Maryland, College Park, MD 20742, USA \\
\email{nathan@\{cs.biu.ac.il; cfar.umd.edu\}}
}

\date{}

\maketitle

\begin{abstract}
In this paper we propose the first effective genetic algorithm (GA)-based jigsaw puzzle solver. We introduce a novel crossover procedure that merges two ``parent'' solutions to an improved ``child'' configuration by detecting, extracting, and combining correctly assembled puzzle segments. The solver proposed exhibits state-of-the-art performance, as far as handling previously attempted puzzles more accurately and efficiently, as well puzzle sizes that have not been attempted before. The extended experimental results provided in this paper include, among others, a thorough inspection of up to 30,745-piece puzzles (compared to previous attempts on 22,755-piece puzzles), using a considerably faster concurrent implementation of the algorithm. Furthermore, we explore the impact of different phases of the novel crossover operator by experimenting with several variants of the GA. Finally, we compare different fitness functions and their effect on the overall results of the GA-based solver.

\end{abstract}

\clearpage

\begin{figure}
\centering
         \begin{subfigure}[t]{0.45\textwidth}
                \centering
                \includegraphics[width=\textwidth]{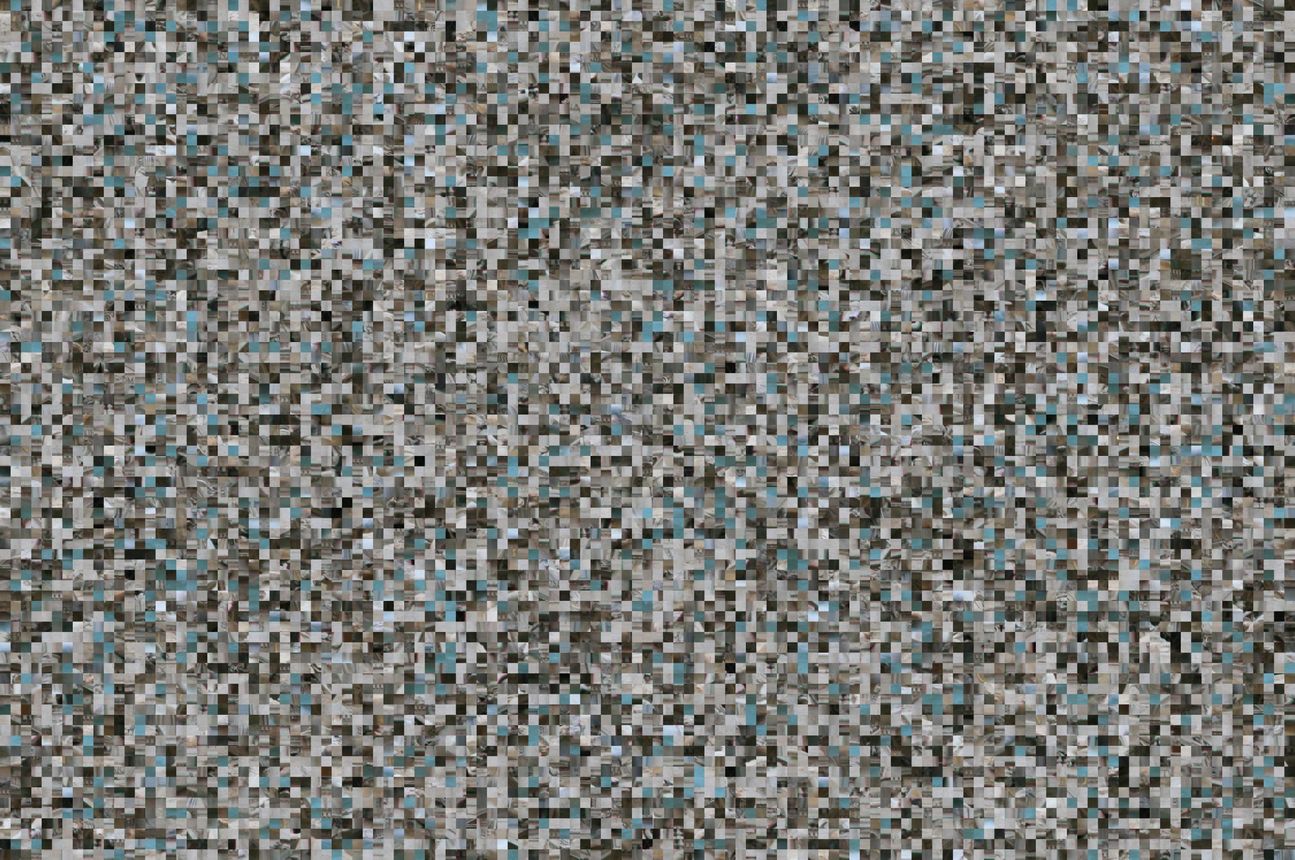}
                \caption{}
                \label{fig:intro_10375_gen_00000000}
        \end{subfigure}
        ~
        \begin{subfigure}[t]{0.45\textwidth}
                \centering
                \includegraphics[width=\textwidth]{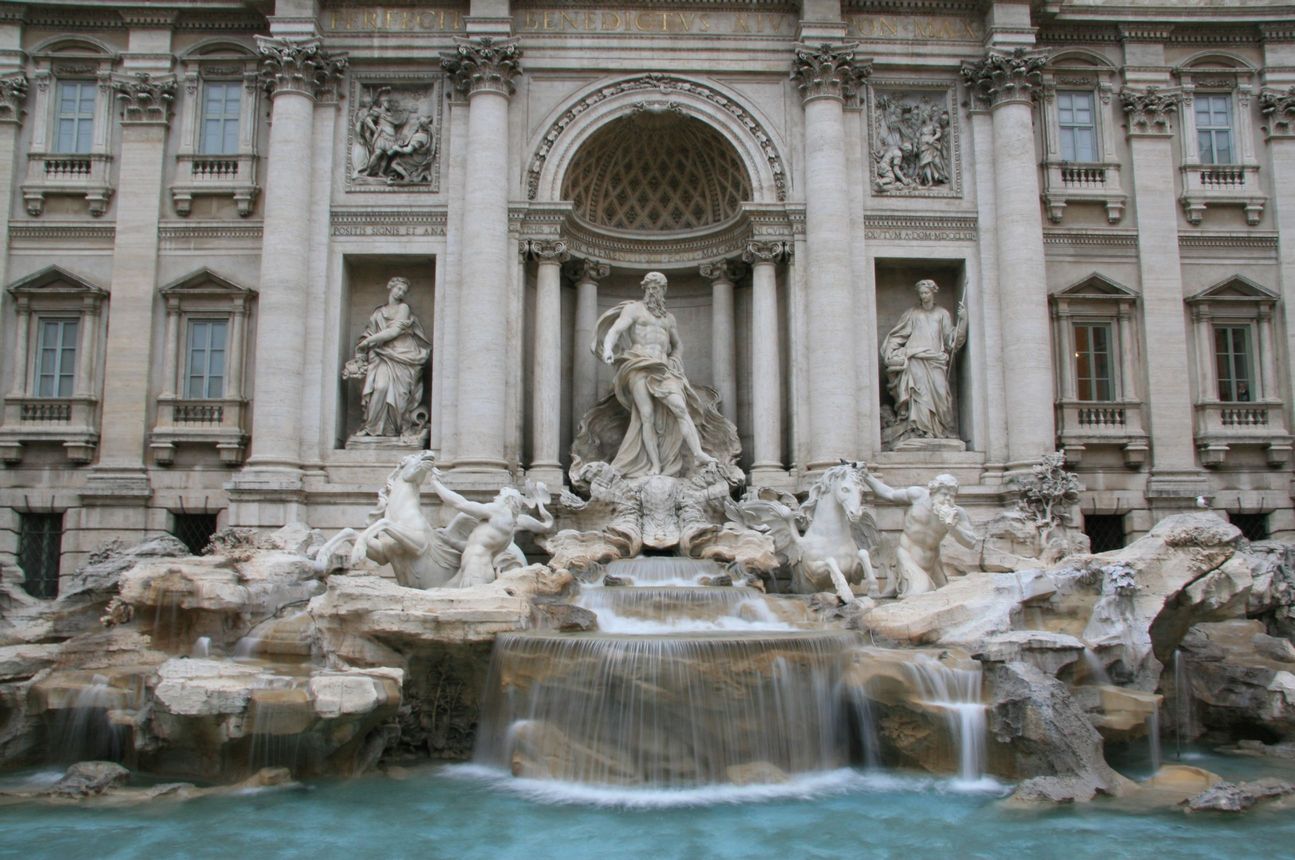}
                \caption{}
                \label{fig:intro_10375_orig}
        \end{subfigure}

	    \begin{subfigure}[t]{0.45\textwidth}
                \centering
                \includegraphics[width=\textwidth]{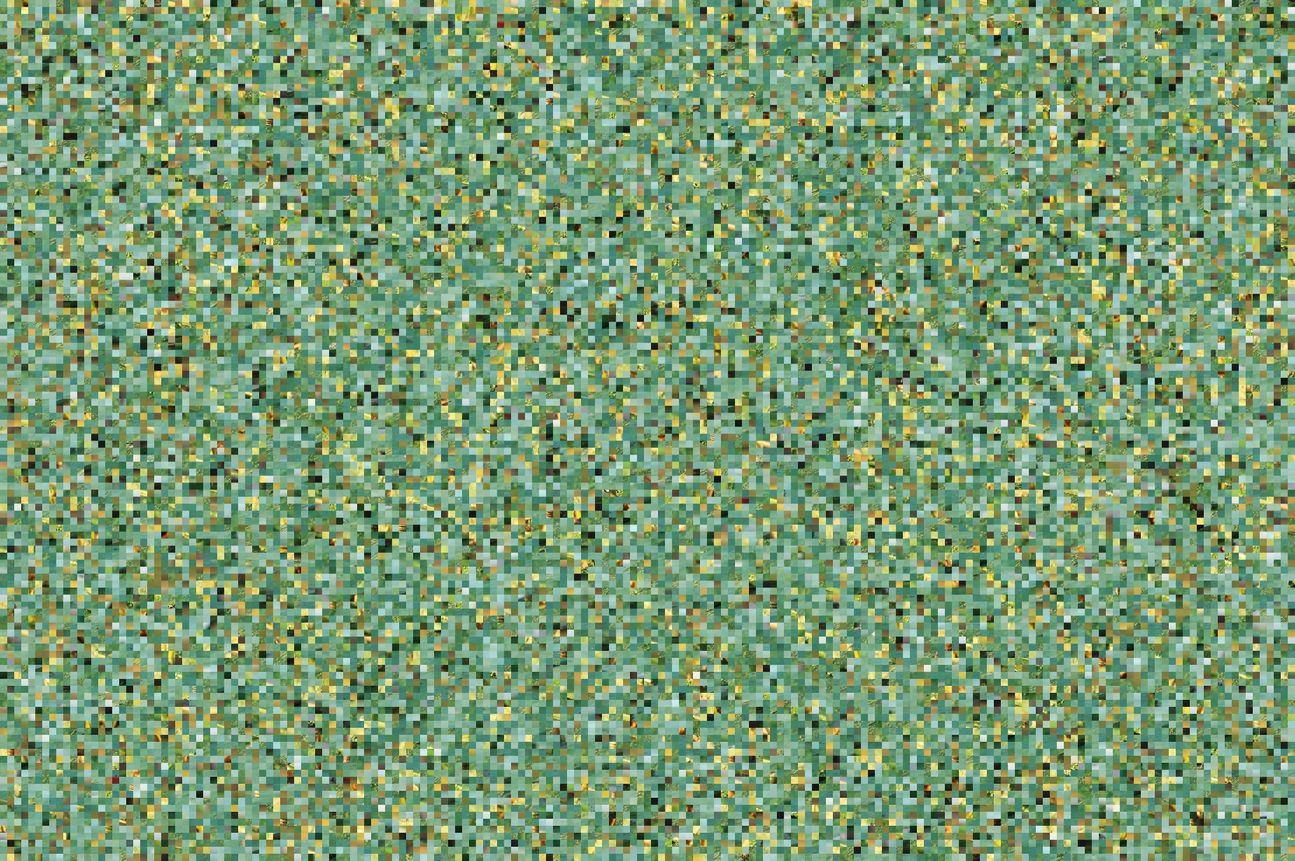}
                \caption{}
                \label{fig:intro_22834_gen_00000000}
        \end{subfigure}%
        ~ 
        \begin{subfigure}[t]{0.45\textwidth}
                \centering
                \includegraphics[width=\textwidth]{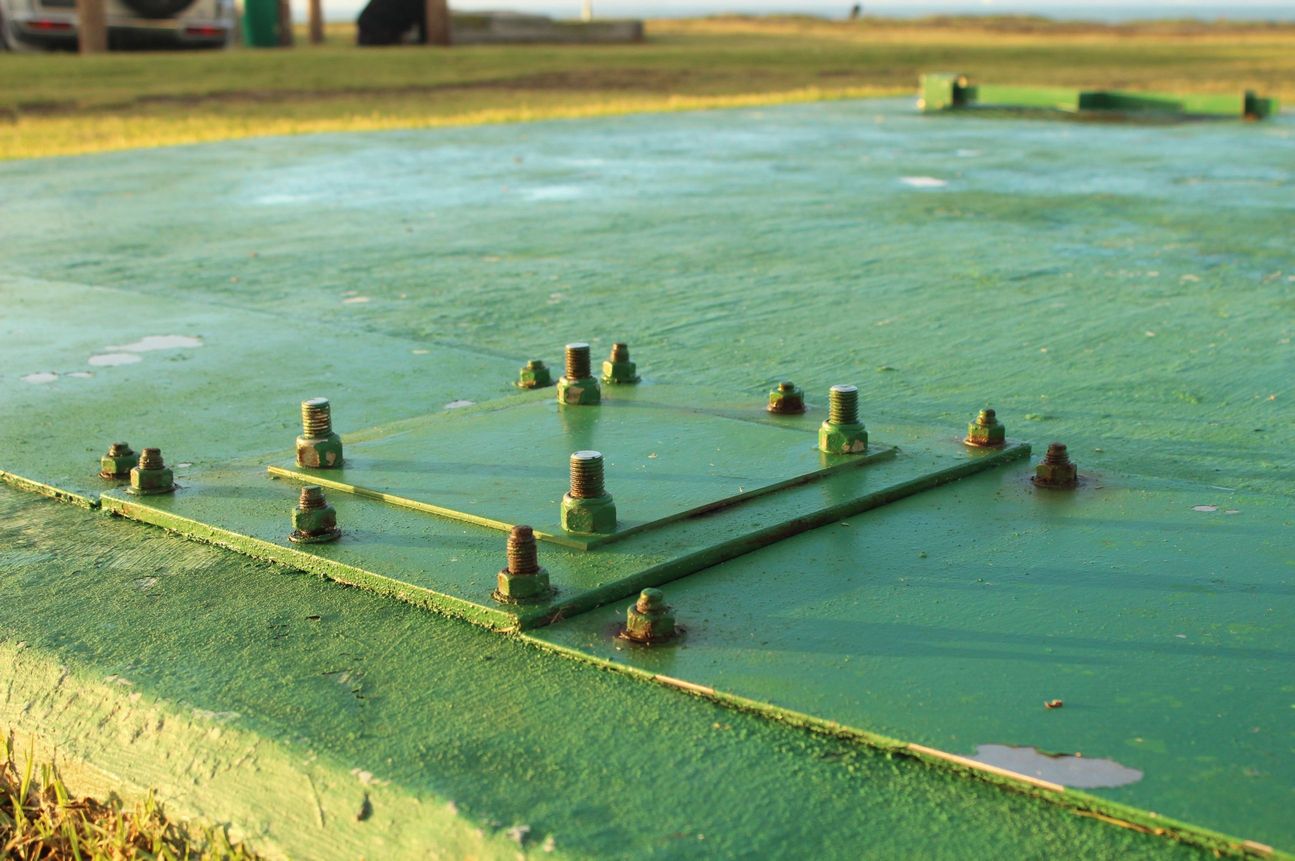}
                \caption{}
                \label{fig:intro_22834_gen_00000100}
        \end{subfigure}
        \caption{ Jigsaw puzzles before and after reassembly using our GA-based solver: (a-b) 10,375- and (c-d) 22,755-piece puzzles, which are among the largest automatically solved puzzles to date. }
        \label{fig:introFig}
\end{figure}

\section{Introduction}
The problem domain of jigsaw puzzles is widely known to almost every human being from childhood. Given $n$ different non-overlapping pieces of an image, the player has to reconstruct the original image, taking advantage of both the shape and chromatic information of each piece. Although this popular game was proven to be an NP-complete problem ~\cite{journals/aai/Altman89,springerlink:10.1007/s00373-007-0713-4}, it has been played successfully by children worldwide. Solutions to this problem might benefit the fields of biology~\cite{journals/science/MarandeB07}, chemistry~\cite{oai:xtcat.oclc.org:OCLCNo/ocm45147791}, literature~\cite{conf/ifip/MortonL68}, speech descrambling~\cite{Zhao:2007:PSA:1348258.1348289}, archeology~\cite{journals/tog/BrownTNBDVDRW08,journals/KollerL06}, image editing~\cite{bb43059} and the recovery of shredded documents or photographs~\cite{cao2010automated,conf/icip/DeeverG12,justino2006reconstructing,marques2009reconstructing}. Besides, as Goldberg {\etal}~\cite{GolMalBer04} have noted, the jigsaw puzzle problem may and should be researched for the sole reason that it stirs pure interest.

Recent years have witnessed a vast improvement in the research and development of automatic jigsaw puzzle solvers, manifested in both puzzle size, solution accuracy, and amount of manual human intervention required. Reported state-of-the-art solvers are fully automated and can handle puzzles of up to 9,000 pieces. Most, if not all, of the aforementioned solvers are greedy, and thus are at great risk of converging to local optima. Despite the great potential in devising a genetic algorithm (GA)-based solver, the success of previous attempts was limited to 64-piece puzzles~\cite{bb58987}, most likely due to the enormous complexity associated with evolutionary computation (EC), in general, and the inherent difficulty of devising an effective crossover operator for this problem, in particular.

Following the convention used in recent work~\cite{conf/cvpr/ChoAF10,conf/cvpr/PomeranzSB11,yang2011particle}, we consider jigsaw puzzles with non-overlapping, ($28 X 28$) square pieces, where both piece orientation and puzzle dimensions are known. (This version, where piece locations are the only unknowns, is called a "Type 1" puzzle~\cite{conf/cvpr/Gallagher12}.) The solver has no knowledge of the original image and may not make any use of it.
In this paper we introduce a novel crossover operator, enabling for the first time an effective GA-based puzzle solver. Our solver compromises neither speed nor size as it outperforms, for the most part, state-of-the-art solvers, tackling successfully up to 30,745-piece size puzzles (i.e., more than three times the number of pieces that has ever been attempted/reported), within a reasonable time frame. (see, {\eg}, Figure~\ref{fig:introFig} for 10,375- and 22,755-piece puzzles.) Our contribution should benefit research regarding EC in general, and the jigsaw puzzle problem, in particular. From an EC perspective, our novel techniques could be used for solving additional problems with similar properties. As to the jigsaw puzzle problem, our proposed framework could prove useful for solving more advanced variants, such as puzzles with missing pieces, unknown piece orientation, mixed puzzles, and more. Finally, we assemble a new benchmark, consisting of sets of larger images (with varying degrees of difficulty), which we make public to the community~\cite{conf/cvpr/site/Our}. Also, we share all of our results (on this benchmark and other public datasets) for future testing and comparative evaluation of jigsaw puzzle solvers.

This paper is an extended version of our previously presented work~\cite{Sholomon_2013_CVPR}. We lay out the requirements from an effective (GA-based) jigsaw puzzle solver, and provide a more detailed description of our crossover operator, as part of our proposed solution. Furthermore, the paper includes new empirical results of an extended set of experiments aimed at a more exhaustive performance evaluation of our presented solver, in general, and its novel crossover operator, in particular. We formed an extended benchmark of up to 30,745-piece puzzles and tested our solver's performance multiple times on each image. Specifically, we have pursued the following empirical issues: (1) Explored the relative impact of the different phases of our 3-phase crossover operator; (2) tested our solver's performance also with a different fitness function; (3) investigated further the shifting anomaly discovered in our early work; (4) introduced a concurrent crossover version to reduce significantly the overall run-time of the GA (without affecting the solver's accuracy).

\section{Previous Work}

Jigsaw puzzles were first introduced around 1760 by John Spilsbury, a Londonian engraver and mapmaker. Nevertheless, the first attempt by the scientific community to computationally solve the problem is attributed to Freeman and Garder~\cite{bb47278} who in 1964 presented a solver which could handle up to nine-piece problems. Ever since then, the research focus regarding the problem has shifted from shape-based to merely color-based solvers of square-tile puzzles. In 2010 Cho {\etal}~\cite{conf/cvpr/ChoAF10} presented a probabilistic puzzle solver that could handle up to 432 pieces, given some a priori knowledge of the puzzle. Their results were improved a year later by Yang {\etal}~\cite{yang2011particle} who presented a particle filter-based solver. Furthermore, Pomeranz {\etal}~\cite{conf/cvpr/PomeranzSB11} introduced that year, for the first time, a fully automated square jigsaw puzzle solver that could handle puzzles of up to 3,000 pieces. Gallagher~\cite{conf/cvpr/Gallagher12} has further advanced this by considering a more general variant of the problem, where neither piece orientation nor puzzle dimensions are known.

In its most basic form, every puzzle solver requires an estimation function to evaluate the compatibility of adjacent pieces and a strategy for placing the pieces as accurately as possible. Although much effort has been invested in perfecting the compatibility functions, recent strategies tend to be greedy, which is known to be problematic when encountering local optima. Thus, despite achieving very good -- if not perfect -- solutions for some puzzles, supplementary material provided by Pomeranz {\etal}~\cite{conf/cvpr/site/PomeranzSB11} suggests there is much room for improvement for many other puzzles. Comparative studies conducted by Gallagher (\cite{conf/cvpr/Gallagher12}, Table 4), regarding the benchmark of 432-piece images, reveal only a slight improvement in accuracy relatively to Pomeranz {\etal} (95.1\% vs. 95.0\%). To the best of our knowledge, no additional runs on other benchmarks have been reported by Gallagher. Interestingly, despite the availability of puzzle solvers for 3,000- and 9,000-piece puzzles, there exists no image set, for the purpose of benchmark testing, containing puzzles with more than 805 pieces. Current state-of-the-art solvers were tested only on very few large images, and those tested contained an extreme variety of textures and colors, which renders them, admittedly, as ``easier'' for solving~\cite{conf/cvpr/Gallagher12}. We assume that as with the smaller images, the accuracy of current solvers on some large puzzles could be greatly improved by using more sophisticated algorithms.

New attempts were made to solve the jigsaw puzzle problem since our original publication~\cite{Sholomon_2013_CVPR}; see, {\eg}~\cite{son2014solving} which is based on the so-called loop constraints. To the best of our knowledge, our GA-based solver exhibits comparable or superior performance (in most cases) to other ``Type 1'' solvers reported in the literature. Furthermore, the proposed GA framework has been adapted successfully to solve considerably harder puzzle variants, {\eg} where the pieces are taken from different puzzles and where the piece orientations and puzzle dimensions are not known. See~\cite{sholomon2014generalized,sholomon2014genetic} for further details.

\section{GA-based puzzle solver}
We use a standard implementation of GA with fitness proportional solution (also known as roulette wheel selection) and uniform crossover. In particular, we use a measure of elitism as in each generation we replicate the best 4 chromosomes. Mutation is embedded inside the crossover operator (which thus might be also referred to as a variation operator). Algorithm~\ref{alg:GAMainLoop} describes the above mentioned framework. The following parameter values are used:
\\
\\
population size = 1000\\
number of generations = 100\\
mutation rate = 0.05 \\

\begin{algorithm}
\caption{Pseudocode of GA Framework}
\label{alg:GAMainLoop}
\begin{algorithmic}[1]
\State {$population \gets $ generate 1000 random chromosomes}
\For{$generation\_number = 1 \to 100$}
    \State {evaluate all chromosomes using the fitness function}
    \State $new\_population \gets NULL$
    \State {copy 4 best chromosomes to $new\_population$}
    \While{$size(new\_population) \leq 1000$}
        \State {$parent1 \gets $ select chromosome}
        \State {$parent2 \gets $ select chromosome}
        \State {$child \gets crossover(parent1, parent2)$}
        \State {add child to $new\_population$}
    \EndWhile
    \State {$population \gets new\_population$}
\EndFor
\end{algorithmic}
\end{algorithm}

Given a puzzle (image) of $(N \times M)$ pieces, we represent a chromosome by an $(N \times M)$ matrix, each entry of which corresponds to a piece number. (A piece is assigned a number according to its initial location in the given puzzle.) Thus, each chromosome represents a complete solution to the jigsaw puzzle problem, {\ie} a suggested placement of all pieces.

Having provided a framework overview, we now describe in greater detail the various critical components of the GA proposed, {\eg} the chromosome representation, fitness function, and crossover operator.

\subsection{The fitness function}
Each chromosome represents a suggested placement of all pieces. The problem variant at hand assumes no knowledge whatsoever of the original image and thus, the correctness of the absolute location of puzzle pieces cannot be estimated in a simple manner. Instead, the pairwise compatibility (defined below) of every pair of adjacent pieces is computed.

A measure ranking the likelihood of two pieces in the original image as adjacent is called {\em compatibility}. Let $C(x_{i}, x_{j}, R)$  denote the compatibility of two puzzle pieces $x_{i}$, $x_{j}$, where $R\in\{l,r,a,b\}$ indicates the spatial relation between these pieces, i.e., whether $x_{j}$ lies, respectively, to the left/right of $x_{i}$, or above or below it.

Cho {\etal}~\cite{conf/cvpr/ChoAF10} explored five possible compatibility measures, of which the {\em dissimilarity measure} of Eq. (\ref{eq:dissimilarity}) was shown to be the most discriminative. Pomeranz {\etal}~\cite{conf/cvpr/PomeranzSB11} further investigated this issue and chose a similar dissimilarity measure with some slight optimizations. The dissimilarity measure relies on the premise that adjacent jigsaw pieces in the original image tend to share similar colors along their abutting edges, {\ie} the sum (over all neighboring pixels) of squared color differences (over all color bands) should be minimal. Assuming pieces $x_{i}$, $x_{j}$ are represented in normalized L*a*b* space by a $K \times K \times 3$ matrix, where $K$ is the height/width of a piece (in pixels), their dissimilarity, where $x_{j}$ is to the right of $x_{i}$, for example, is
\begin{equation} \label{eq:dissimilarity}
D(x_{i},x_{j},r)=\sqrt{\sum_{k=1}^{K}\sum_{cb=1}^{3}(x_{i}(k,K,cb)-x_{j}(k,1,cb))^{2}},
\end{equation}
where $cb$ stands for color band. It is important to note that dissimilarity is not a metric, since almost always $D(x_{i},x_{j},R) {\neq} D(x_{j},x_{i},R)$. Obviously, to maximize the compatibility of two pieces, their dissimilarity should be minimized.

Another important consideration in choosing a fitness function is that of run-time cost. Since every chromosome in every generation must be evaluated, a fitness function must be relatively computationally-inexpensive. We chose  the standard dissimilarity, as it meets this criterion and also seems to be sufficiently discriminative. To speed up further the computation of the fitness function we added a lookup table of size $2 \cdot (N \cdot M)^2$ containing all of the pairwise compatibilities for all pieces (we need keep only compatibilities with respect to ``right'' and ``above'', as those with respect to ``left'' and ``below'' can be easily deduced).

Finally, the fitness function of a given chromosome is the sum of pairwise dissimilarities over all neighboring pieces (whose configuration is represented by the chromosome). Representing a chromosome by an $(N \times M)$ matrix, where a matrix entry $x_{i,j} (i = 1..N, j = 1..M)$ corresponds to a single puzzle piece, we define its fitness as
\begin{equation} \label{eq:fitness}
\sum_{i=1}^{N}\sum_{j=1}^{M-1}(D(x_{i,j},x_{i,j+1},r))+\sum_{i=1}^{N-1}\sum_{j=1}^{M}(D(x_{i,j},x_{i+1,j},b))
\end{equation}
where, as previously indicated, $r$ and $b$ stand for ``right'' and ``below'', respectively.

\subsection{The crossover operator}
\label{sec:repAndCross}

\subsubsection{Problem definition}

As noted above, a chromosome is represented by an $(N \times M)$ matrix, where each matrix entry corresponds to a puzzle piece number. This representation is straightforward and lends itself easily to the evaluation of the fitness function described above. The main issue concerning this representation is the design of an appropriate crossover operator. A naive crossover operator with respect to the given representation will create a new child chromosome at random, such that each entry of the resulting matrix is the corresponding cell of the first or second parent. This approach yields usually a child chromosome with duplicate and/or missing puzzle pieces, which is of course an invalid solution to the problem. The inherent difficulty surrounding the crossover operator may well have played a critical role in delaying thus far the development of a state-of-the-art solution to the problem, based on evolutionary computation.

Once the validity issue is rectified, one needs to consider very carefully the crossover operator. Recall, crossover is applied to two chromosomes selected due to their high fitness values, where the fitness function used is an overall pairwise compatibility measure of adjacent puzzle pieces. At best, the function rewards a correct placement of neighboring pieces next to each other, but it has no way of identifying the actual correct location of a piece. Since the population starts out from a random piece placement and then improves gradually, it is reasonable to assume that some correctly assembled puzzle segments emerge over the generations. Taking into account the fitness function's inability to reward a correct position, we expect such segments to appear most likely at incorrect locations. A crossover operator should pass on ``good traits'' from the parents to the child, thereby creating possibly a better solution. Discovering a correct segment is not trivial; it should be regarded as a good trait that needs to be exploited. Namely, it should not be discarded, but rather trickled to the next generations. Thus, the crossover operator must allow for {\em position independence}, {\ie} the ability of shifting entire correctly-assembled segments, in an attempt to place them correctly in their true location in the child chromosome.

Once the position-independence issue is taken care of, one should address the issue of detecting correctly-assembled segments, which are possibly misplaced. What segment should the crossover operator pass on to an offspring? A random approach might seem appealing, but in practice it could be infeasible due to the enormous size of the solution domain of the problem. Some heuristics may be applied to distinguish correct segments from incorrect ones.

In summary, a good crossover operator should address the issues of validity of child chromosomes, detection of supposedly correctly-assembled puzzle segments in parents, and position independence of these segments when passed on to an offspring.

\subsubsection{Our proposed solution}

\begin{figure*}
\centering
        \begin{subfigure}[t]{0.30\textwidth}
                \centering
                \includegraphics[width=\textwidth]{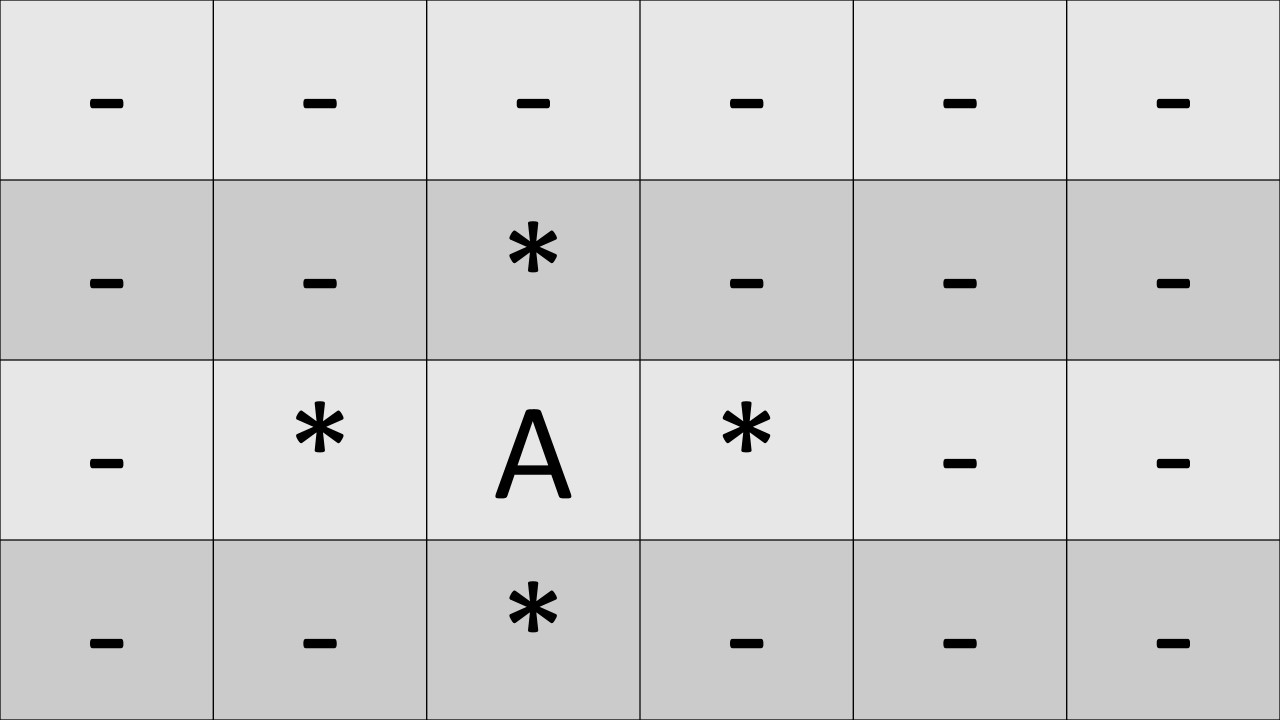}
                \caption{}
                \label{fig:result_10375_04_gen_00000001}
        \end{subfigure}
        ~ 
        \begin{subfigure}[t]{0.30\textwidth}
                \centering
                \includegraphics[width=\textwidth]{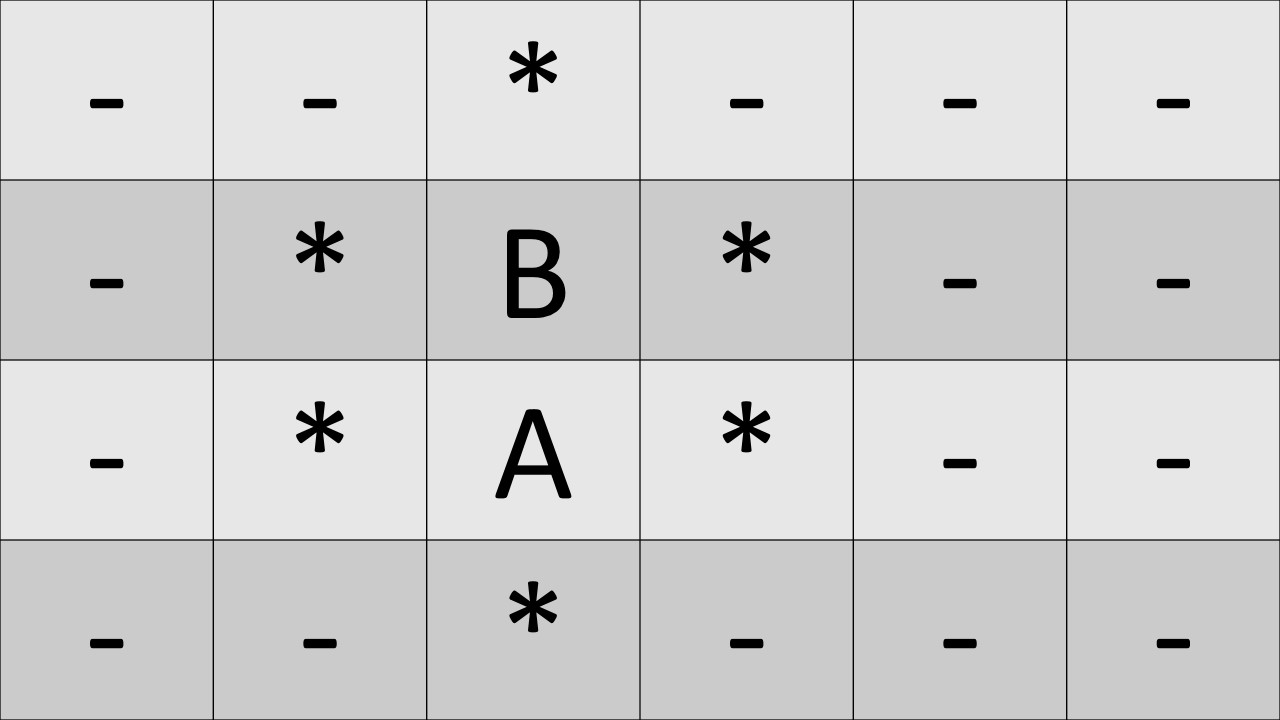}
                \caption{}
                \label{fig:result_10375_04_gen_00000002}
        \end{subfigure}
        ~
        \begin{subfigure}[t]{0.30\textwidth}
                \centering
                \includegraphics[width=\textwidth]{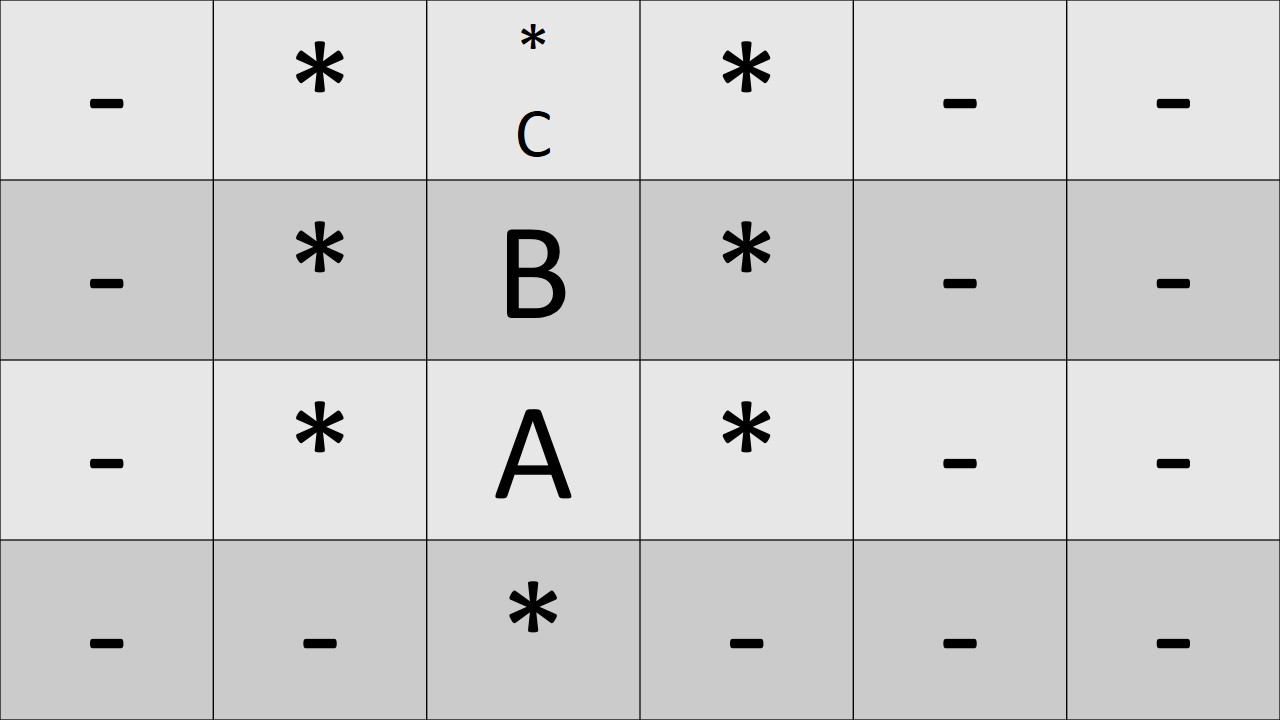}
                \caption{}
                \label{fig:result_10375_04_gen_00000100}
        \end{subfigure}
        ~
        \begin{subfigure}[t]{0.30\textwidth}
                \centering
                \includegraphics[width=\textwidth]{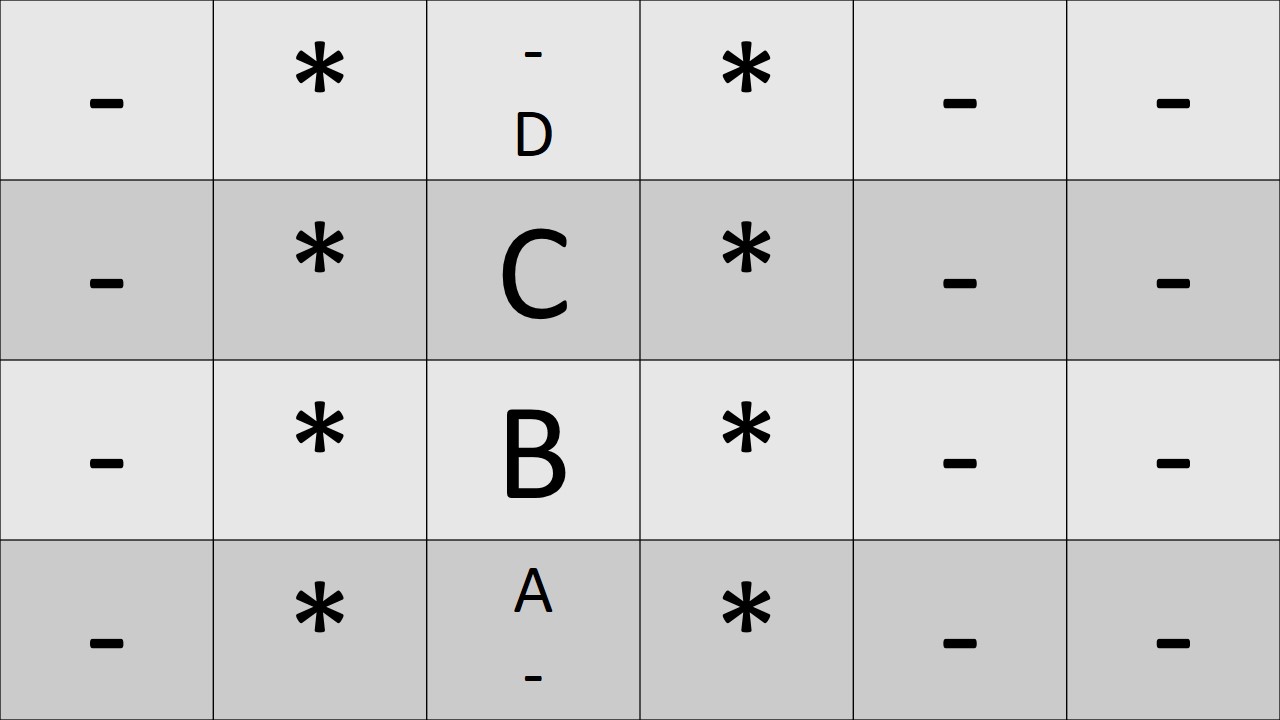}
                \caption{}
                \label{fig:mgcShifted_down}
        \end{subfigure}
        ~
        \begin{subfigure}[t]{0.30\textwidth}
                \centering
                \includegraphics[width=\textwidth]{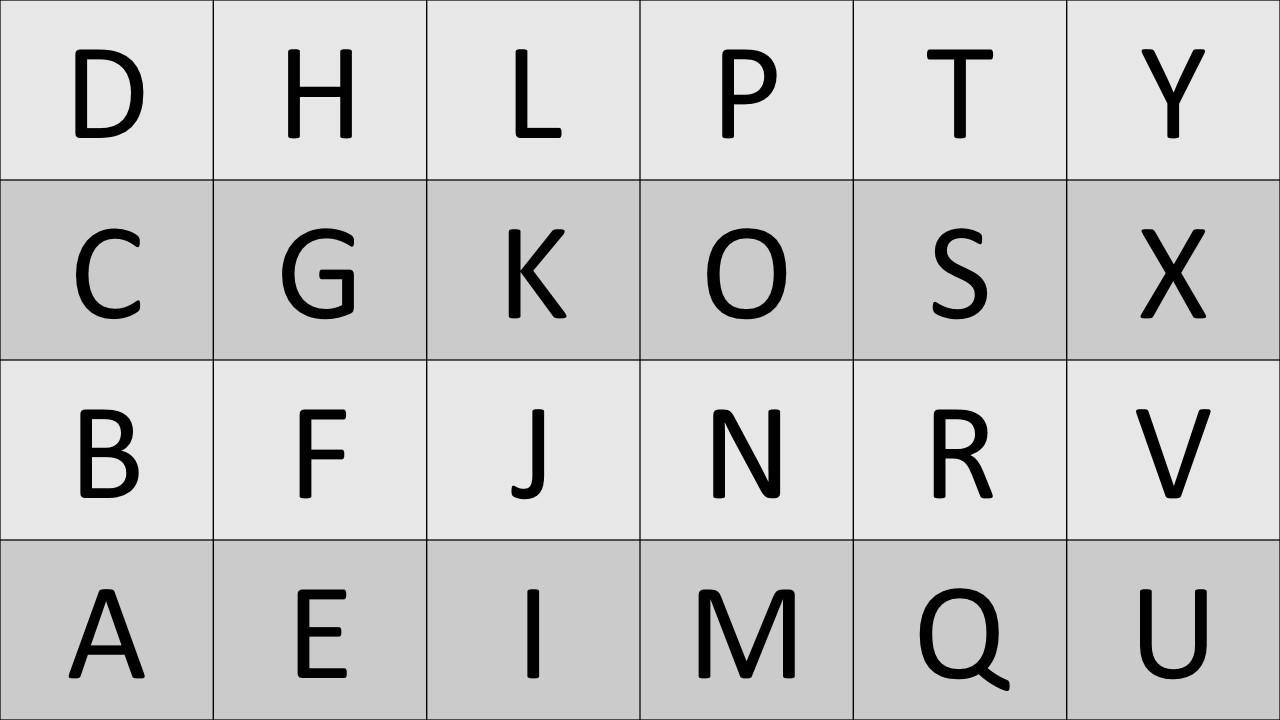}
                \caption{Resulting Image}
                \label{fig:result_10375_04_gen_00000000}
        \end{subfigure}%
        \caption{Illustration of kernel growing; puzzle pieces are depicted by letters. Slot marked by ``*'' can be assigned a piece by  crossover operator; slot marked by ``-'' cannot, since it is not adjacent to any assigned piece: (a)  Kernel's first stage, with only a single piece, (b)--(d) assignment of next three pieces; note shift downward of entire kernel in (d) , {\ie} piece $A$ does not end up where it is placed initially, and (e) resulting image, where entire column in (d) was shifted to the left. }
        \label{fig:ABCKernel}
\end{figure*}

Given two parent chromosomes, {\ie} two complete different arrangements of all puzzle pieces, the crossover operator constructs a child chromosome in a kernel-growing fashion, using both parents as ``consultants''. The operator starts with a single piece and gradually joins other pieces at available kernel boundaries. New pieces may be joined only adjacently to existing ones, so that the emerging image is always contiguous. For example, let $A$ be the first piece chosen by the operator. The second piece $B$ must be placed in a neighboring slot, i.e., left/right of, above or below piece $A$. Assuming it is placed just above $A$, then a third piece $C$ must be assigned to one of the empty, previously mentioned slots (i.e., left/right of, or below $A$), or to any of the newly available slots that are left/right of, or above $B$. See Figure~\ref{fig:ABCKernel} for an illustration of the above scenario.

\begin{figure*}
\centering
        \begin{subfigure}[t]{0.30\textwidth}
                \centering
                \includegraphics[width=\textwidth]{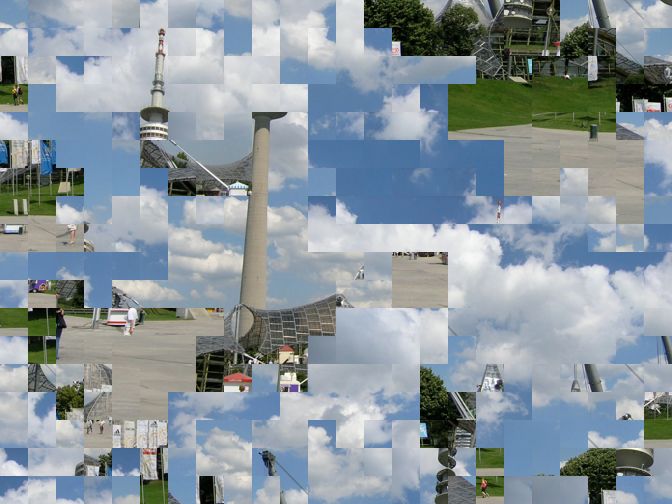}
                \caption{Parent1}
                \label{fig:result_10375_04_gen_00000000}
        \end{subfigure}%
        ~ 
        \begin{subfigure}[t]{0.30\textwidth}
                \centering
                \includegraphics[width=\textwidth]{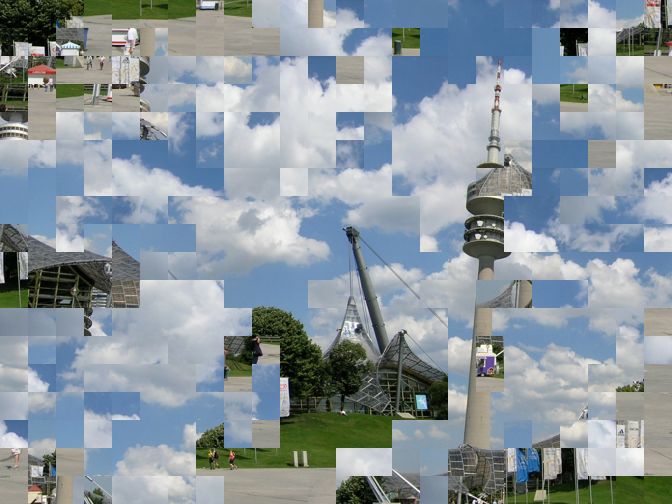}
                \caption{Parent2}
                \label{fig:result_10375_04_gen_00000001}
        \end{subfigure}
        ~ 
        \begin{subfigure}[t]{0.30\textwidth}
                \centering
                \includegraphics[width=\textwidth]{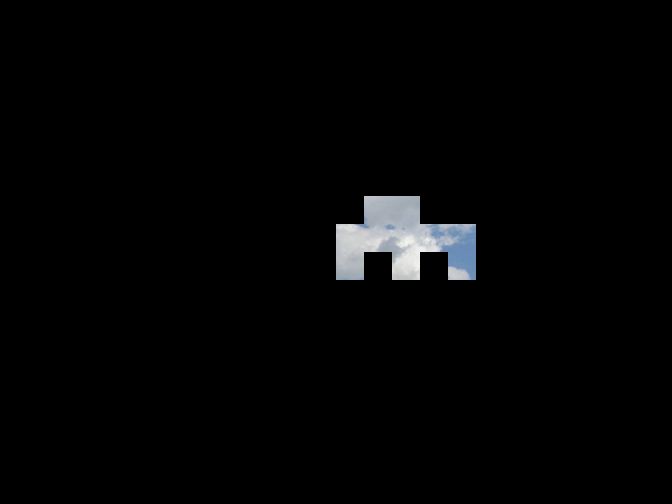}
                \caption{10 pieces}
                \label{fig:result_10375_04_gen_00000002}
        \end{subfigure}
        ~
        \begin{subfigure}[t]{0.30\textwidth}
                \centering
                \includegraphics[width=\textwidth]{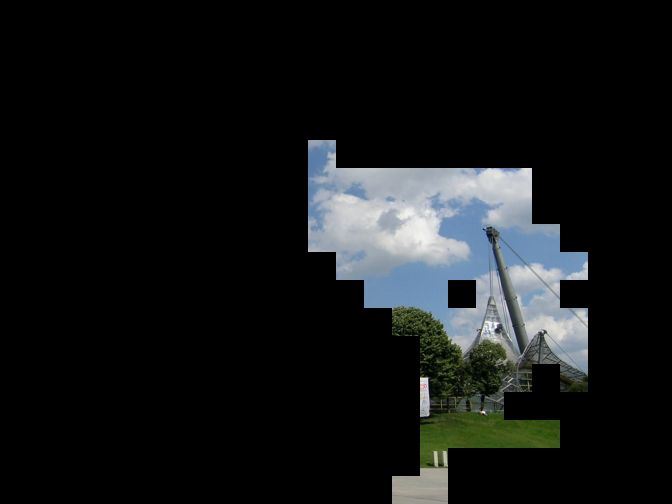}
                \caption{70 pieces}
                \label{fig:result_10375_04_gen_00000002}
        \end{subfigure}
        ~
        \begin{subfigure}[t]{0.30\textwidth}
                \centering
                \includegraphics[width=\textwidth]{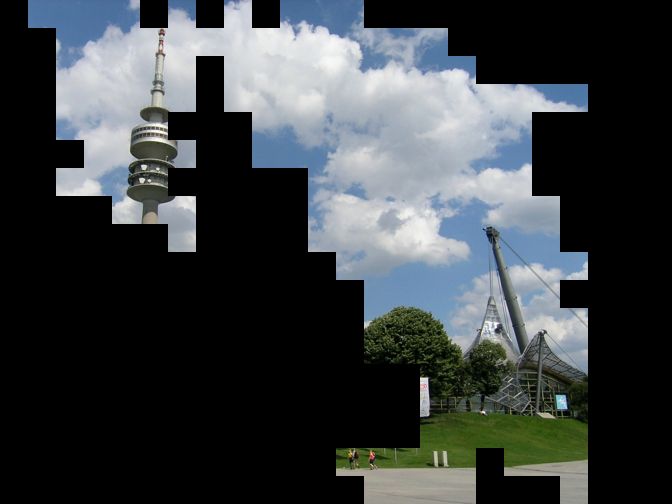}
                \caption{180 pieces}
                \label{fig:result_10375_04_gen_00000002}
        \end{subfigure}
        ~
        \begin{subfigure}[t]{0.30\textwidth}
                \centering
                \includegraphics[width=\textwidth]{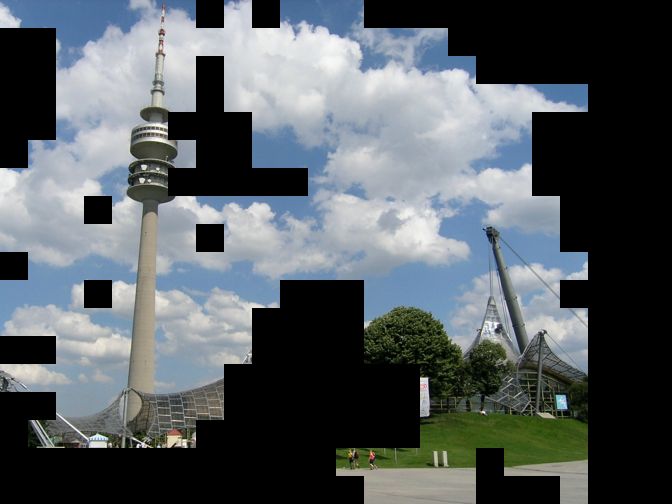}
                \caption{258 pieces}
                \label{fig:result_10375_04_gen_00000002}
        \end{subfigure}
        ~
        \begin{subfigure}[t]{0.30\textwidth}
                \centering
                \includegraphics[width=\textwidth]{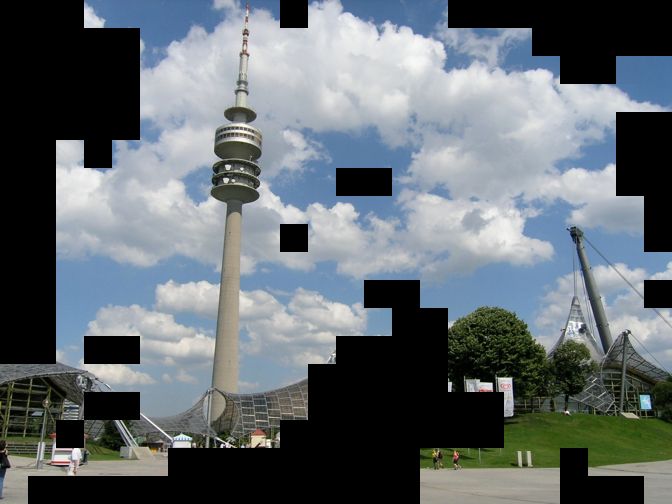}
                \caption{304 pieces}
                \label{fig:result_10375_04_gen_00000002}
        \end{subfigure}
        ~
        \begin{subfigure}[t]{0.30\textwidth}
                \centering
                \includegraphics[width=\textwidth]{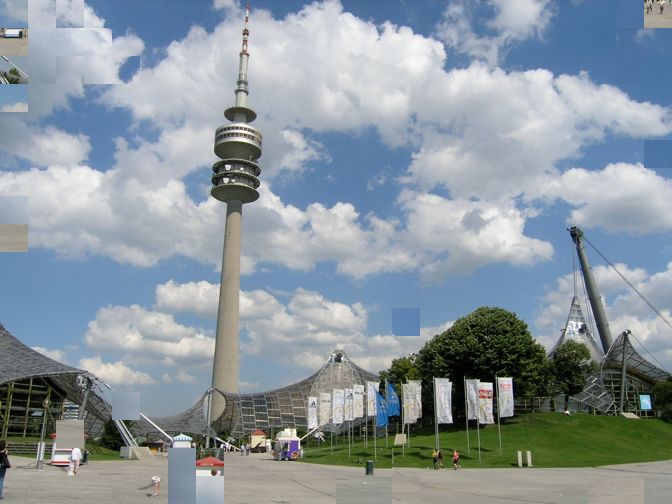}
                \caption{Child}
                \label{fig:result_10375_04_gen_00000100}
        \end{subfigure}
        \caption{ Illustration of crossover operation: Given (a) Parent1 and (b) Parent2, (c) -- (g) depict gradual growth of kernel of pieces until (h) complete child obtained; note the detection of tower parts in both parents, which are shifted and merged into complete tower; image shifting during kernel growing is due to so-called piece-position independence. }
        \label{fig:growingKernel}
\end{figure*}

The operator keeps adding tiles from a bank of available puzzle pieces until no pieces are left. Hence, every piece will appear exactly once in the resulting image. Since the image size is known in advance, the operator can ensure that there is no boundary violation. Thus, by using every piece exactly once inside of a frame of the correct size, the operator is guaranteed of achieving a valid image. Figure~\ref{fig:growingKernel} illustrates the above described kernel-growing process.

A key property of the kernel-growing technique is that the actual final location of every piece is only determined once the kernel reaches its final size and the child chromosome is complete. Before that, all pieces might be shifted, depending on the kernel's growth vector. The first piece, for example, might end up at the lower-left corner of the image, if the kernel should grow only upwards and to the right, after this piece is assigned. On the other hand, the very same piece might be placed ultimately at the center of the image, its upper-right corner, or any other location, for that matter. This change in a tile's actual location is illustrated in Figures~\ref{fig:ABCKernel} and ~\ref{fig:growingKernel}; see, in particular, Figure~\ref{fig:growingKernel}(f)--(g) of the kernel-growing process, showing pieces shifted to the right due to the insertion of new pieces on the left. It is this important property which enables the position independence of image segments. Namely, a correctly-assembled puzzle segment in a parent chromosome, possibly misplaced with respect to its true location, could be copied over during the crossover operation (while preserving its structure) to its correct location in an offspring.

We described how to construct a child chromosome as a growing kernel. The remaining issues are: (1) Which piece to select from the available bank of tiles and (2) where to place it in the child ({\ie} at which available neighboring slot). Given a kernel, {\ie} a partial image, we can mark all the boundaries where a new piece might be placed. A piece boundary is denoted by a pair $(x_{i}, R)$, consisting of the piece number and a spatial relation. The operator invokes a three-phase procedure. First, given all existing boundaries, the operator checks whether there exists a piece boundary which both parents agree on, {\ie} whether there exists a piece $x_{j}$ that is in the spatial direction $R$ of $x_{i}$) in both parents. If so, then $x_{j}$ is added to the kernel in the appropriate location. If the parents agree on two or more boundaries, one of them is chosen at random and the corresponding piece is assigned. Obviously, if a piece is already in use, it cannot be (re)assigned, {\ie} it is ignored as if the parents do not agree on that particular boundary.

If there is no agreement between the parents on any piece at any boundary, the second phase begins. To understand this phase, we briefly review the concept of a {\em best-buddy} piece, first introduced by Pomeranz {\etal}~\cite{conf/cvpr/PomeranzSB11}. Two pieces are said to be best-buddies if each piece considers the other as its most compatible piece. Pieces $x_{i}$ and $x_{j}$ are said to best-buddies if
\begin{align}
\forall x_{k} \in Pieces, \; C(x_{i},x_{j},R_1) \geq C(x_{i},x_{k},R_1)\notag \\
\text{and \quad\quad\quad\quad\quad\quad\quad\quad}\\
\forall x_{p} \in Pieces, \; C(x_{j},x_{i},R_2) \geq C(x_{j},x_{p},R_2) \notag
\end{align}
where $Pieces$ is the set of all given image pieces and $R_1$ and $R_2$ are ``complementary'' spatial relations ({\eg} if $R_1$ = right, then $R_2$ = left and vice versa). In the second phase, the operator checks whether one of the parents contains a piece $x_j$ in spatial relation $R$ of $x_i$ which is also a best-buddy of $x_i$ with respect to that relation. If so, the piece is chosen and assigned. As before, if multiple best-buddy pieces are available, one of them is chosen at random. If a best-buddy piece is found that is already assigned, it is ignored and the search continues for other best-buddy pieces. If no best-buddy piece exists, the operator proceeds to the final third phase, where it selects a boundary at random and assigns to it the most compatible piece available. To introduce mutation, the operator places, with low probability, in the first and last phase, an available piece at random, instead of the most compatible relevant piece available.

In summary, the operator uses repeatedly a three-phase procedure of piece selection and assignment, first placing agreed-upon pieces, followed by best-buddy pieces, and finally by the most compatible piece available ({\ie} the most compatible piece not yet assigned). An assignment is only considered at relevant boundaries to maintain the contiguity of the kernel-growing image. The procedure returns to the first phase after every piece assignment due to the creation of new prospective boundaries. Algorithm~\ref{alg:CrossoverSimplified} provides a simplified description of the crossover operator (without mutation).

\begin{algorithm}
\caption{Crossover operator simplified}
\label{alg:CrossoverSimplified}
\begin{algorithmic}[1]
\State {If any available boundary meets the criterion of Phase 1 (both parents agree on a piece), place the piece there and goto (1); otherwise continue.}
\State {If any available boundary meets the criterion of Phase 2 (one parent contains a best-buddy piece), place the piece there and goto (1); otherwise continue.}
\State {Randomly choose a boundary, place the most compatible available piece there and goto (1).}
\end{algorithmic}
\end{algorithm}

\subsubsection{Rationale}

We expect to see child chromosomes inherit ``good'' traits from their parent chromosomes in an effective GA framework. Since the algorithm encourages piece position independence, the trait of interest is captured by sets of neighboring pieces, rather than the locations of particular pieces. Correct puzzle segments correspond to the correct placement of pieces relatively to each other. The notion, for example, that piece $x_{i}$ is in spatial relation $R$ to piece $x_{j}$  is key to solving the jigsaw puzzle problem. Thus, although every chromosome accounts for a complete placement of all the pieces, it is the internal relative piece assignments that are examined and exploited.

We assume that a trait common to both parents under consideration has propagated through the generations and is essentially the reason for their survival and selection. In other words, if both parents agree on a spatial relation, the algorithm would consider it real, and would attempt to draw on it, with high probability. Given, however, the highly randomized nature of the first generations, it might prove counterproductive that agreed-upon pieces selected initially would persist through the latter generations. Due to the kernel-growing algorithm, some agreed-upon pieces might be selected as most compatible pieces at other boundaries and thus be discarded for later use. For example, let pieces $A$ and $B$ be adjacent in both parents; still, piece $B$ might be selected during the kernel growing and placed next to a different piece $C$. In other words, not all agreed-upon relations are necessarily copied over to the child. Thus, random agreements in early generations could be nullified.

As for the second stage, where the parents agree on no piece, the algorithm could randomly pick a parent and follow its lead. Another possibility might be to just choose the most compatible piece in a greedy manner, or check if a best-buddy piece is available. Since piece assignment in the parents might have been random, and since even best-buddy pieces might not capture the correct match, we combine these two elements. The fact that two pieces are both best-buddies and are adjacent in a parent is a good indication for the validity of this match. A different approach is to assume that every chromosome contains some correct segments. The transfer of correct segments from parents to children is at the heart of the GA. Moreover, if two parents contain a correct segment and these segments overlap, the common overlapping part will be copied over to the child in the first phase and be completed from both parents in the second phase. This would result in combining the segments into a larger correct segment, obtaining an enhanced child chromosome, and advancing the overall correct solution.

As for the greedy third step, the GA concurrently tries many different greedy placements; only those that seem correct propagate through the generations. This exemplifies the principle of propagation of good traits in the spirit of the theory of natural selection.

\section{Experimental results}
\begin{figure*}
\centering
        \begin{subfigure}[t]{0.45\textwidth}
                \centering
                \includegraphics[width=\textwidth]{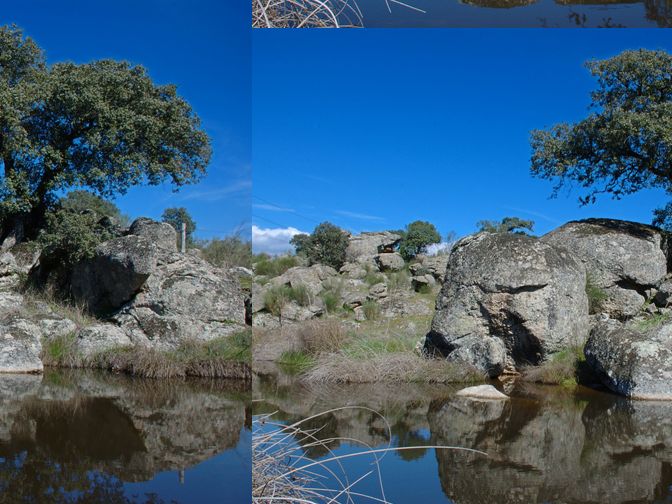}
                \caption{}
                \label{fig:result_10375_04_gen_00000000}
        \end{subfigure}%
        ~ 
        \begin{subfigure}[t]{0.45\textwidth}
                \centering
                \includegraphics[width=\textwidth]{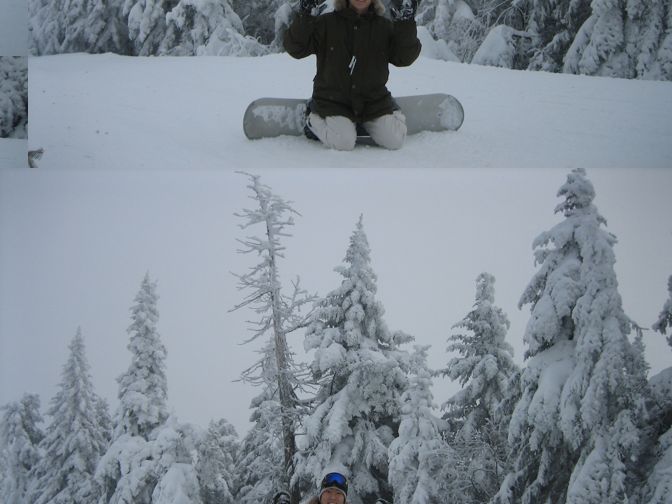}
                \caption{}
                \label{fig:result_10375_04_gen_00000001}
        \end{subfigure}
        ~ 
        \begin{subfigure}[t]{0.45\textwidth}
                \centering
                \includegraphics[width=\textwidth]{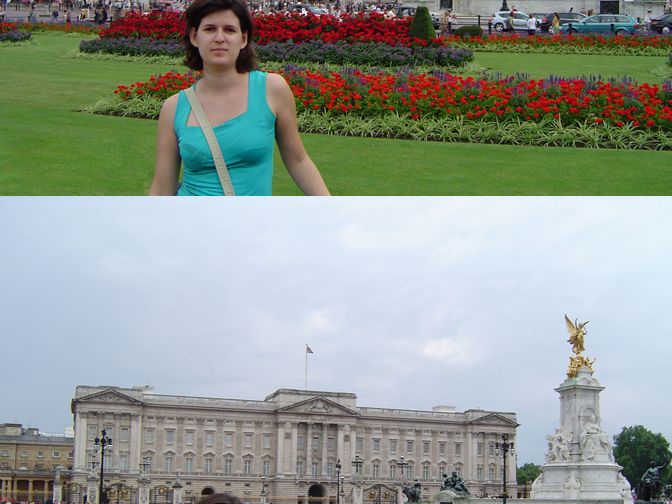}
                \caption{}
                \label{fig:result_10375_04_gen_00000002}
        \end{subfigure}
        ~
        \begin{subfigure}[t]{0.45\textwidth}
                \centering
                \includegraphics[width=\textwidth]{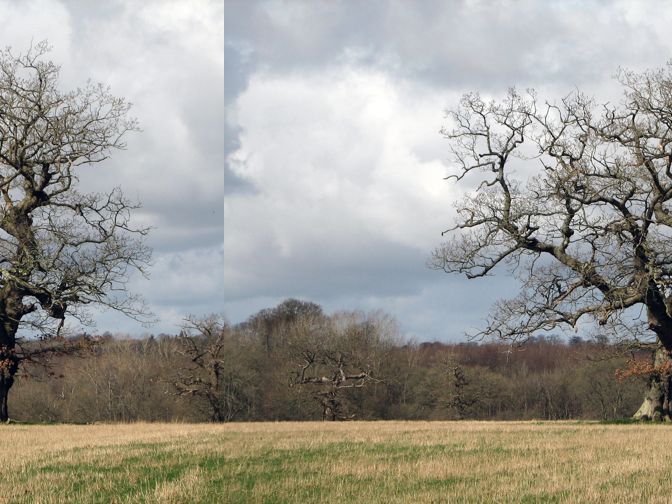}
                \caption{}
                \label{fig:result_10375_04_gen_00000100}
        \end{subfigure}
        \caption{(a)?(d): Shifted solutions created by our GA; the accuracy for each solution is 0\% according to the direct comparison, but over 95\% according to the (more meaningful) neighbor comparison; amazingly, the dissimilarity of each solution is smaller than that of its original image counterpart. }
        \label{fig:shifted}
\end{figure*}

Cho {\etal}~\cite{conf/cvpr/ChoAF10} introduced the following two main measures, which have been repeatedly used in previous works, to evaluate the accuracy of an assembled puzzle: The {\em direct comparison}, which measures the fraction of pieces located in their correct location, and the {\em neighbor comparison}, which measures the fraction of correct neighbors. The direct method is considered ~\cite{conf/cvpr/PomeranzSB11} less accurate and less meaningful due to its inadequate evaluation of slightly shifted puzzle solutions. Figure~\ref{fig:shifted} illustrates the drawbacks of the direct comparison and the superiority of the neighbor comparison. Note that a piece configuration scoring 100\% according to one definition implies the complete reconstruction of the original image and will also achieve a perfect score according to the other definition. Thus, unless stated otherwise, all results are presented with respect to the neighbor comparison. (For the sake of completeness, our results with respect to the direct comparison are reported in Table~\ref{tab:directAvg}.)

\begin{figure*}
\centering
      \begin{subfigure}[t]{0.23\textwidth}
                \centering
                \includegraphics[width=\textwidth]{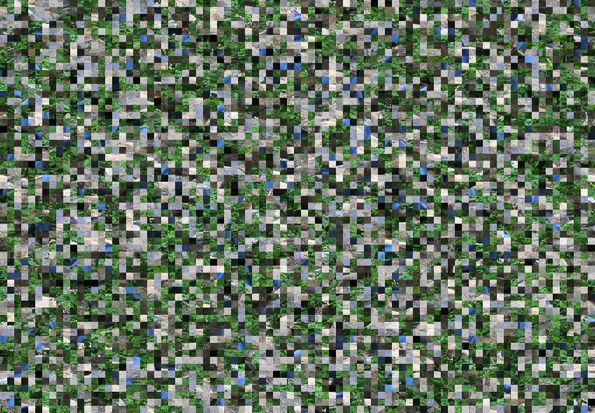}
                \caption{5,015 pieces}
                \label{fig:result_5015_19_gen_00000000}
        \end{subfigure}%
        ~ 
        \begin{subfigure}[t]{0.23\textwidth}
                \centering
                \includegraphics[width=\textwidth]{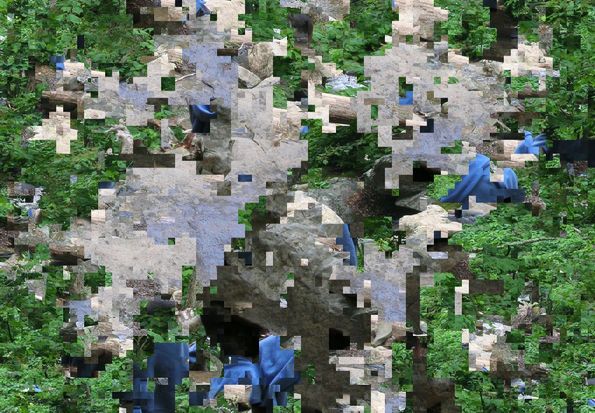}
                \caption{Generation 1}
                \label{fig:result_5015_19_gen_00000001}
        \end{subfigure}
        ~ 
        \begin{subfigure}[t]{0.23\textwidth}
                \centering
                \includegraphics[width=\textwidth]{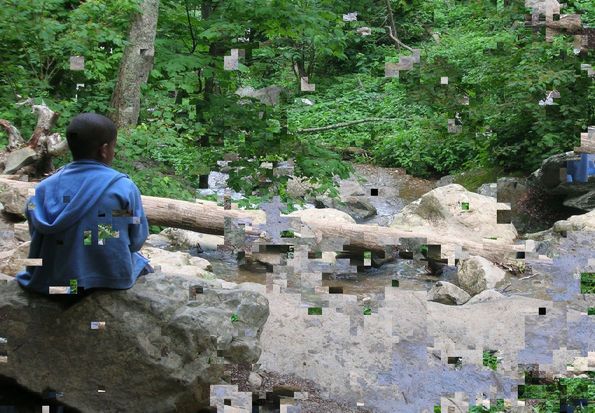}
                \caption{Generation 2}
                \label{fig:result_5015_19_gen_00000002}
        \end{subfigure}
        ~
        \begin{subfigure}[t]{0.23\textwidth}
                \centering
                \includegraphics[width=\textwidth]{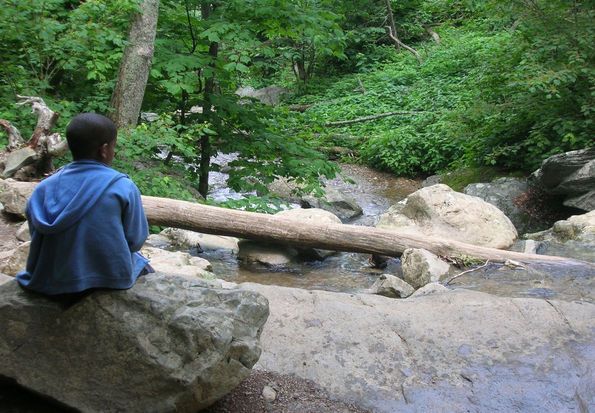}
                \caption{Final}
                \label{fig:result_5015_19_gen_00000100}
        \end{subfigure}

        \begin{subfigure}[t]{0.23\textwidth}
                \centering
                \includegraphics[width=\textwidth]{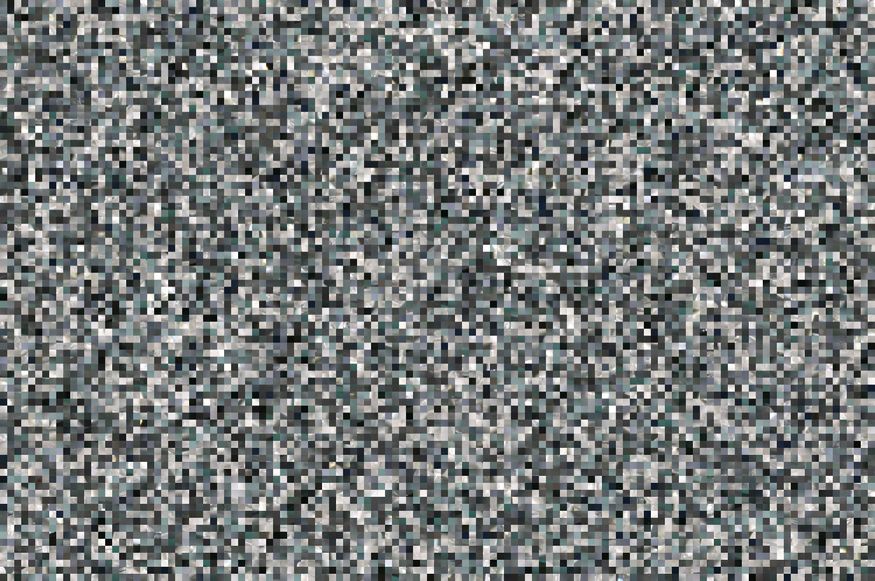}
                \caption{10,375 pieces}
                \label{fig:result_10375_04_gen_00000000}
        \end{subfigure}%
        ~ 
        \begin{subfigure}[t]{0.23\textwidth}
                \centering
                \includegraphics[width=\textwidth]{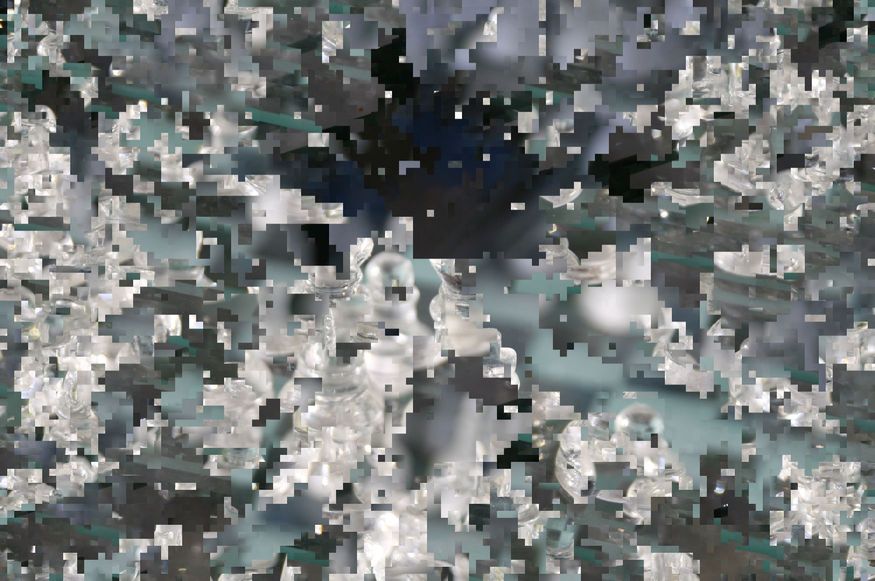}
                \caption{Generation 1}
                \label{fig:result_10375_04_gen_00000001}
        \end{subfigure}
        ~ 
        \begin{subfigure}[t]{0.23\textwidth}
                \centering
                \includegraphics[width=\textwidth]{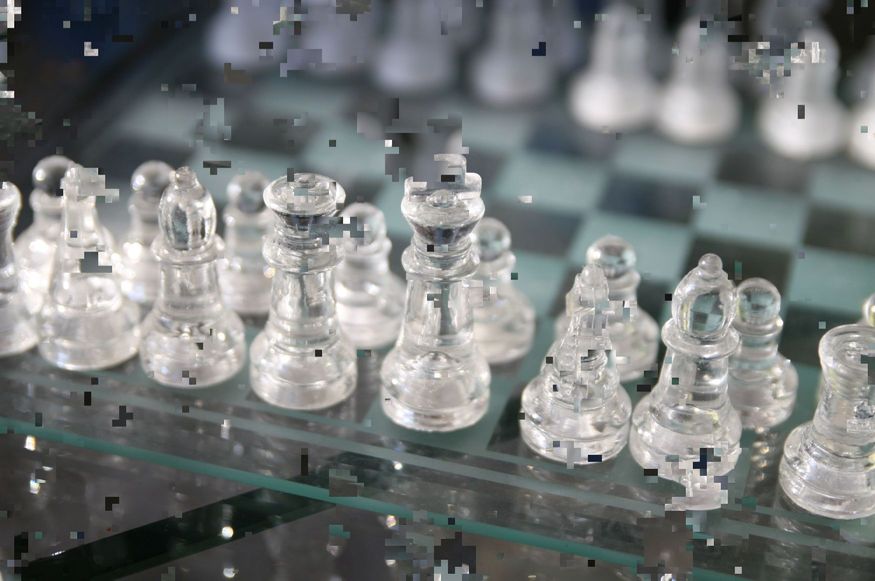}
                \caption{Generation 2}
                \label{fig:result_10375_04_gen_00000002}
        \end{subfigure}
        ~
        \begin{subfigure}[t]{0.23\textwidth}
                \centering
                \includegraphics[width=\textwidth]{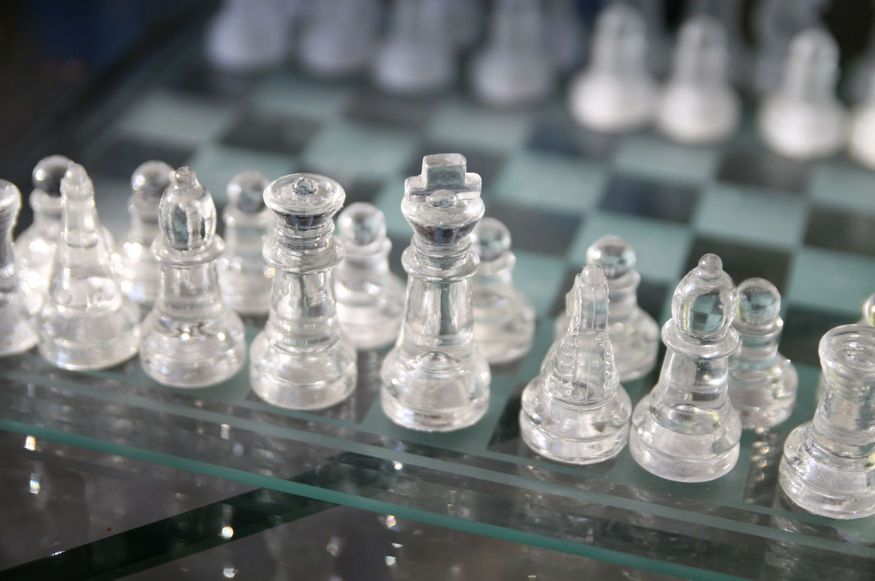}
                \caption{Final}
                \label{fig:result_10375_04_gen_00000100}
        \end{subfigure}

        \begin{subfigure}[t]{0.23\textwidth}
                \centering
                \includegraphics[width=\textwidth]{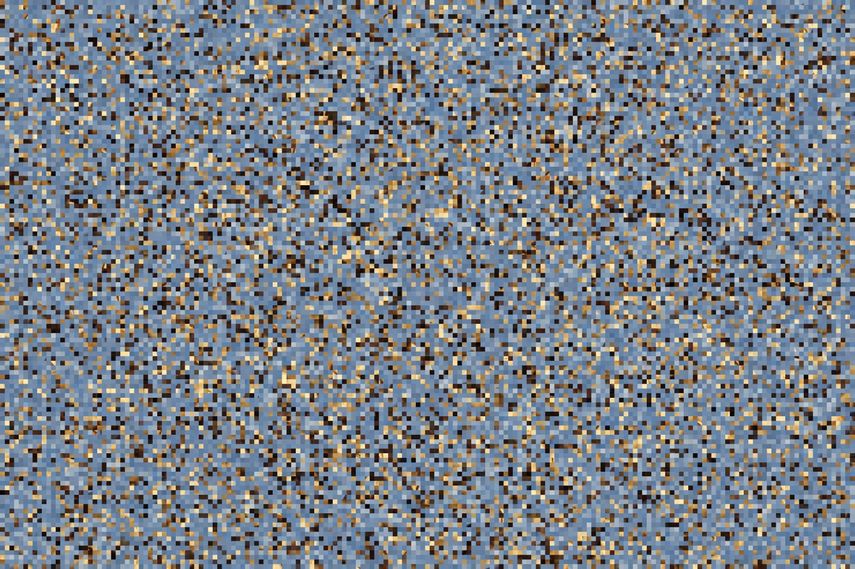}
                \caption{22,755 pieces}
                \label{fig:result_22834_12_gen_00000000}
        \end{subfigure}%
        ~ 
        \begin{subfigure}[t]{0.23\textwidth}
                \centering
                \includegraphics[width=\textwidth]{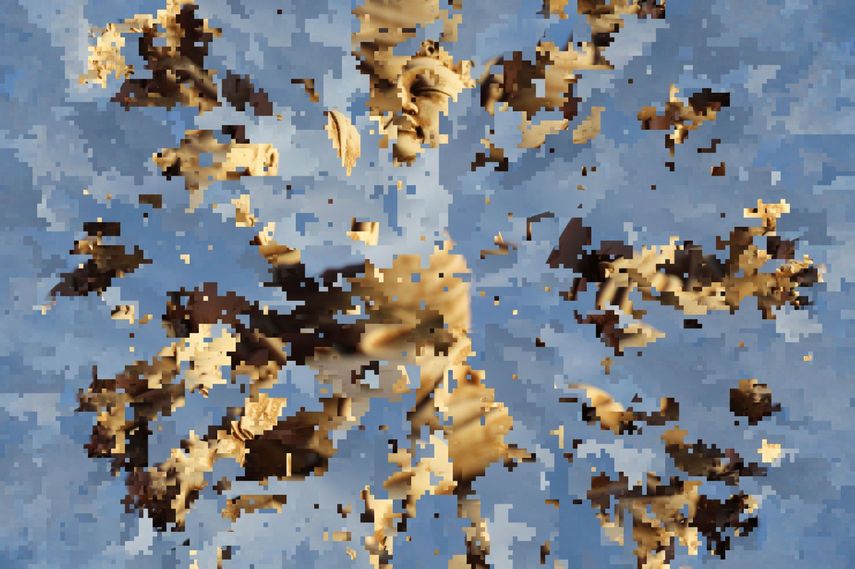}
                \caption{Generation 1}
                \label{fig:result_22834_12_gen_00000001}
        \end{subfigure}
        ~ 
        \begin{subfigure}[t]{0.23\textwidth}
                \centering
                \includegraphics[width=\textwidth]{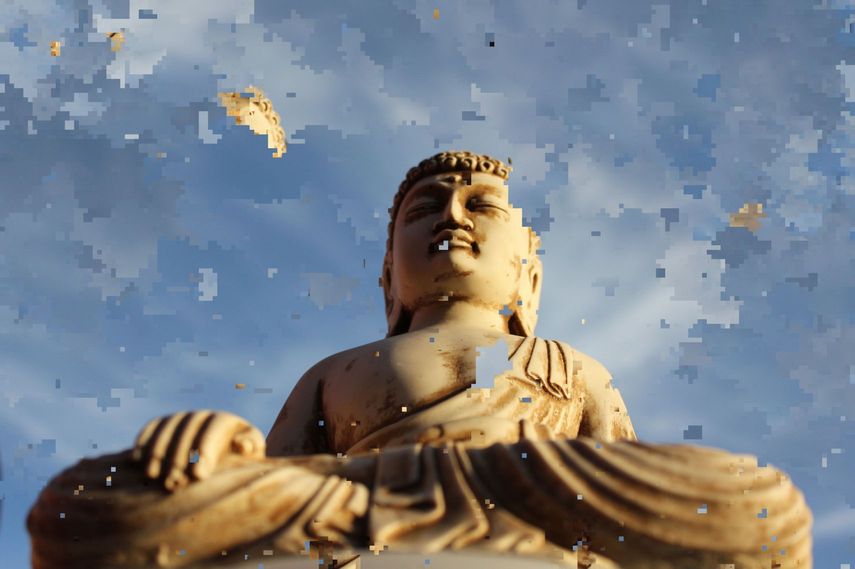}
                \caption{Generation 2}
                \label{fig:result_22834_12_gen_00000002}
        \end{subfigure}
        ~
        \begin{subfigure}[t]{0.23\textwidth}
                \centering
                \includegraphics[width=\textwidth]{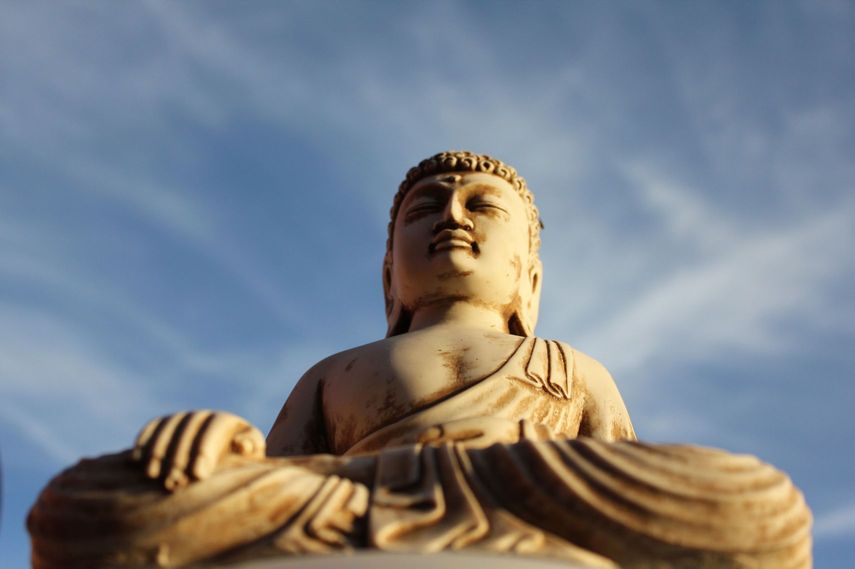}
                \caption{Final}
                \label{fig:result_22834_12_gen_00000007_final}
        \end{subfigure}

        \begin{subfigure}[t]{0.23\textwidth}
                \centering
                \includegraphics[width=\textwidth]{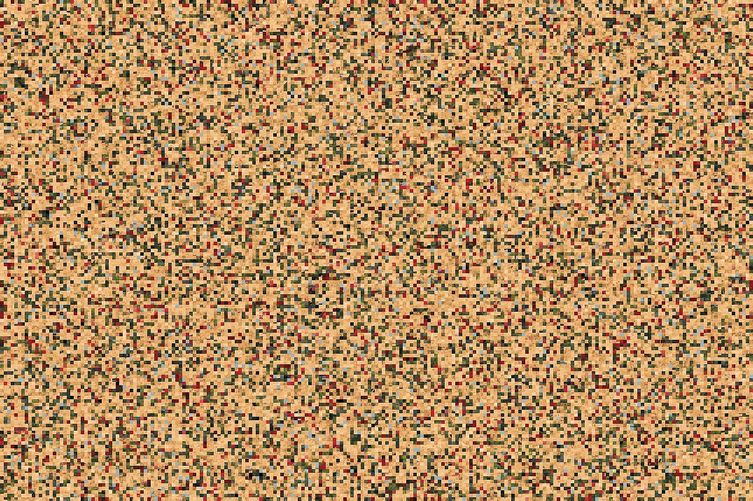}
                \caption{30,745 pieces}
                \label{fig:result_22834_12_gen_00000000}
        \end{subfigure}%
        ~ 
        \begin{subfigure}[t]{0.23\textwidth}
                \centering
                \includegraphics[width=\textwidth]{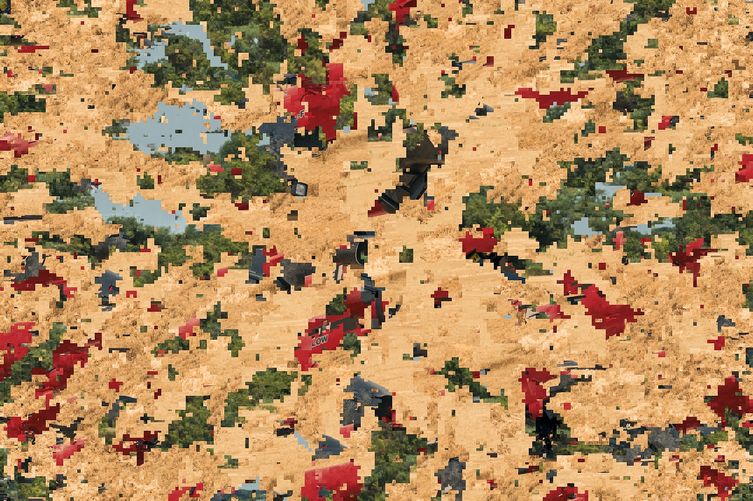}
                \caption{Generation 1}
                \label{fig:result_22834_12_gen_00000001}
        \end{subfigure}
        ~ 
        \begin{subfigure}[t]{0.23\textwidth}
                \centering
                \includegraphics[width=\textwidth]{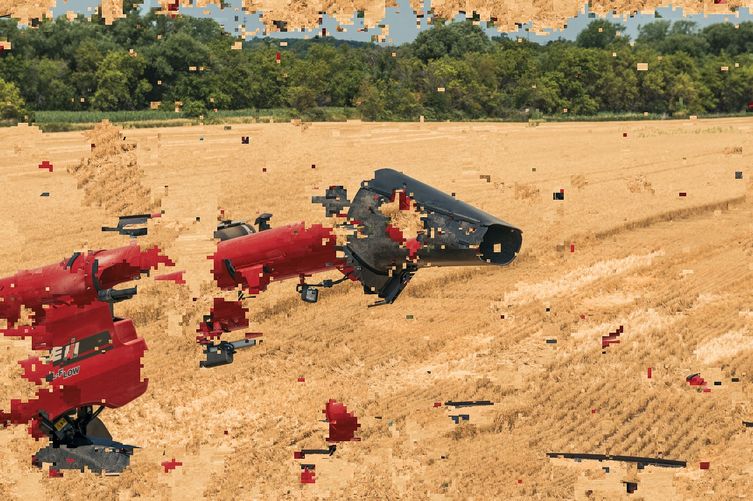}
                \caption{Generation 2}
                \label{fig:result_22834_12_gen_00000002}
        \end{subfigure}
        ~
        \begin{subfigure}[t]{0.23\textwidth}
                \centering
                \includegraphics[width=\textwidth]{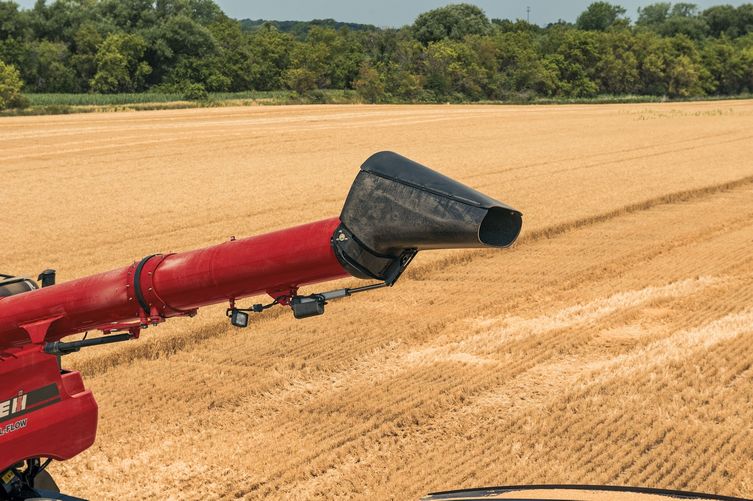}
                \caption{Final}
                \label{fig:result_22834_12_gen_00000007_final}
        \end{subfigure}

        \caption{Selected results of our GA-based solver for large puzzles. The first row shows: (a) 5,015-piece puzzle and the best chromosome achieved by the GA in the (b) first, (c) second, and (d) last generation. Similarly, the second and third rows show the best chromosomes in the same generations for 10,375-, 22,755- and 30,745-piece puzzles, respectively. The accuracy of all puzzle solutions shown is 100\%.}
        \label{fig:selectedSolutions}
\end{figure*}

We first tested the proposed GA on the set of images supplied by Cho {\etal}~\cite{conf/cvpr/ChoAF10} and all sets supplied by Pomeranz {\etal}~\cite{conf/cvpr/PomeranzSB11}, testing puzzles of 28 $\times$ 28-pixel patches according to the traditional convention. The image data experimented with contain 20-image sets of 432-, 504-, and 805-piece puzzles and 3-image sets of 2,360- and 3,360-piece puzzles. Next, we have augmented the above benchmark by compiling four additional 20-image sets for this work and future studies. These additional sets contain images of 5,015-, 10,375-, 22,755 and 30,745-piece puzzles. (Note that the latter two puzzle sizes have not been attempted before.) We ran the GA ten times on each image (see Figure~\ref{fig:selectedSolutions} for selected results), every time with a different random seed, and recorded ?- for these ten runs ?- the best, worst, and average accuracy (as well as the standard deviation). The averages of these best, worst, and average results are referred to as ``average best'', ``average worst'', and ``average average'', respectively. Tables~\ref{tab:directAvg} and~\ref{tab:bestWorstAvgComp} list the results achieved by our GA on each set for the direct comparison and the neighbor comparison, respectively. Interestingly, despite the inherent randomness of GAs, different runs yield almost identical results, attesting to the solver's robustness. Having observed little difference between the best and worst runs for the largest size puzzles, in particular, we settled for only five GA runs for each image of the 22,755- and 30,745-piece puzzles. (Note that the GA achieves very high accuracy also on these very large image sizes that have not been attempted before.)

Table~\ref{tab:pomAvgCompare} compares -- for each image set -- our average best results to those of Pomeranz {\etal}, which can be easily derived from their well-documented results~\cite{conf/cvpr/site/PomeranzSB11}. As can be seen, our GA results are more accurate than theirs. Nevertheless, as noted in the beginning of this paper, previous solvers might perform well on some images and worse on others. The advantageous performance of our solver is further demonstrated in Table~\ref{tab:pomAvgWorseCompare}, relating only to the three least accurately solved images in every set. Our solver gains a significant improvement of up to 21\% (for the 540-piece puzzle set); for some puzzles, the improvement was even 30\%. In Table~\ref{tab:CompareAll} we provide a detailed comparison of our results, for 432- to 30K-piece puzzles, with those obtained by Pomeranz {\etal}~\cite{conf/cvpr/PomeranzSB11}, Gallagher~\cite{conf/cvpr/Gallagher12}, and Son {\etal}~\cite{son2014solving}, based on the principle of so-called loop constraints~\cite{son2014solving}. The comprehensive set of results demonstrates that the proposed GA remains a state-of-the-art solver. Detailed results of the exact accuracy of every run on every image can be found in our supplementary material~\cite{conf/cvpr/site/Our}.

\begin{table}[t]
\centering
\begin{tabular}{ |c||c|c|c||c| }
    \hline
  \# & Avg. & Avg. & Avg. & Avg. \\
  of Pieces  & Best & Worst & Avg. & Std. Dev. \\ \hline \hline
  432 & 86.19\% & 80.56\% & 82.94\% & 2.62\% \\ \hline
  540 & 92.75\% & 90.57\% & 91.65\% & 0.65\% \\ \hline
  805 & 94.67\% & 92.79\% & 93.63\% & 0.62\% \\ \hline
  2,360 & 85.73\% & 82.73\% & 84.62\% & 0.86\% \\ \hline
  3,300 & 89.92\% & 65.42\% & 86.62\% & 7.19\% \\ \hline
  5,015 & 94.78\% & 90.76\% & 92.04\% & 1.74\% \\ \hline
  10,375 & 97.69\% & 96.08\% & 97.12\% & 0.45\% \\ \hline
  22,755 & 93.41\% & 87.04\% & 90.49\% & 2.59\% \\ \hline
  30,745 & 90.76\% & 81.19\% & 86.11\% & 4.51\% \\ \hline

\end{tabular}
\caption{Results of running the GA ten times on every image in every set under direct comparison; the best, worst, and average scores were recorded for every image.}
\label{tab:directAvg}
\end{table}

\begin{table}[t]
\centering
\begin{tabular}{ |c||c|c|c||c| }
    \hline
  \# & Avg. & Avg. & Avg. & Avg. \\
  of Pieces  & Best & Worst & Avg. & Std. Dev. \\ \hline \hline
  432 & 96.16\% & 95.21\% & 95.70\% & 0.34\% \\ \hline
  540 & 95.96\% & 94.65\% & 95.38\% & 0.40\% \\ \hline
  805 & 96.26\% & 95.35\% & 95.85\% & 0.31\% \\ \hline
  2,360 & 88.86\% & 87.52\% & 88.00\% & 0.38\% \\ \hline
  3,300 & 92.76\% & 91.91\% & 92.37\% & 0.27\% \\ \hline
  5,015 & 95.25\% & 94.87\% & 95.06\% & 0.11\% \\ \hline
  10,375 & 98.47\% & 98.20\% & 98.36\% & 0.08\% \\ \hline
  22,755 & 96.40\% & 95.97\% & 96.19\% & 0.12\% \\ \hline
  30,745 & 93.40\% & 93.12\% & 93.27\% & 0.11\% \\ \hline
\end{tabular}
\caption{Accuracy results using our GA. Averages of best, worst, and average scores (as well as standard deviations) are given for each image set under neighbor comparison; GA is invoked five times on each image of 22,755- and 30,745-piece puzzles and ten times on all other images.}
\label{tab:bestWorstAvgComp}
\end{table}

\begin{table}[t]
\centering
\begin{tabular}{ |c||c|c||c| }
    \hline
  \# of Pieces  & Pomeranz {\etal} & GA & Diff \\ \hline \hline
  432 & 94.25\% & 96.16\% & {\bf 1.91\%} \\ \hline
  540 & 90.90\% & 95.96\%  & {\bf 5.06\%} \\ \hline
  805 & 89.70\% & 96.26\% & {\bf 6.56\%} \\ \hline
  2,360 & 84.67\% & 88.86\% & {\bf 4.19\%} \\ \hline
  3,300 & 85.00\% & 92.76\% & {\bf 7.76\%} \\ \hline
\end{tabular}
\caption{Comparison of our accuracy results to those of Pomeranz {\etal} (derived
from their supplementary material); averages are given for each image set of best
scores over ten runs for each image. }
\label{tab:pomAvgCompare}
\end{table}

\begin{table}[t]
\centering
\begin{tabular}{ |c||c|c||c| }
    \hline
  \# of Pieces  & Pomeranz {\etal} & GA & Diff \\ \hline \hline
  432 & 76.00\% & 81.06\% & {\bf 5.06\%} \\ \hline
  540 & 58.33\% & 79.32\% & {\bf 20.99\%} \\ \hline
  805 & 67.33\% & 86.30\% & {\bf 18.97\%} \\ \hline
  2,360 & 84.67\% & 88.86\% & {\bf 4.19\%} \\ \hline
  3,300 & 85.00\% & 92.76\% & {\bf 7.76\%} \\ \hline
\end{tabular}
\caption{Comparison of our accuracy results to those of Pomeranz {\etal}, relating only to the three least accurately solved images in every set.}
\label{tab:pomAvgWorseCompare}
\end{table}

\begin{table}[t]
\centering
\begin{tabular}{ |c||c|c|c|c| }
    \hline
  \# of Pieces  & Pomeranz {\etal} & Gallagher & Son {\etal} & GA \\ \hline \hline
  432 & 94.25\% & 95.10\% & 95.50\% & {\bf 96.16\%} \\ \hline
  540 & 90.90\% &  - & 95.20\% & {\bf 95.96\%} \\ \hline
  805 & 89.70\% &  - & 94.90\% & {\bf 96.26\%} \\ \hline
  2,360 & 84.67\% &  - & {\bf 96.40\%} & 88.86\% \\ \hline
  3,300 & 85.00\% &  - & {\bf 96.40\%} & 92.76\% \\ \hline
  5,015 & - & - & -& {\bf 95.25\%} \\ \hline
  10,375 & - & - & -& {\bf 98.47\%} \\ \hline
  22,755 & - & - & -& {\bf 96.40\%} \\ \hline
  30,745 & - & - & -& {\bf 93.40\%} \\ \hline
\end{tabular}
\caption{Comparison of our accuracy results to those of Pomeranz {\etal}~\cite{conf/cvpr/PomeranzSB11}, Gallagher~\cite{conf/cvpr/Gallagher12} and Son {\etal}~\cite{son2014solving}. }
\label{tab:CompareAll}
\end{table}

An interesting phenomenon observed is depicted in Figure~\ref{fig:shifted}. For some puzzles, the GA obtains a ``better-than-perfect'' score, {\ie} it returns a configuration for which the dissimilarity is smaller than that of the original (correct) image. Such tile (mis)placements were reproduced even with more sophisticated metrics such as the compatibility measure proposed in ~\cite{conf/cvpr/PomeranzSB11}. Moreover, in some cases, these abnormal solutions are arrived at after the GA does find  the correct solution. As far as we know, observations of this kind have not been reported before. Although undesirable, this phenomenon attests to the GA's capability of optimizing the overall cost function for a given image, thereby avoiding potentially local optima. Obviously, revisiting the fitness function ({\ie} compatibility measure) would be required in an attempt to handle such outcomes.

Finally, Table~\ref{tab:timesTable} depicts the average run time of the GA per image set; all experiments were conducted on a modern PC. When running on the smaller sets of 432, 540, and 805 pieces, the GA terminated after 48.73, 64.06, and 116.18 seconds, on average, respectively. These results are comparable to the running times reported by Pomeranz {\etal}~\cite{conf/cvpr/PomeranzSB11}, {\ie} 1.2, 1.9, and 5.1 minutes, respectively, on the same sets. Experimenting with the larger benchmark of 22,755-piece puzzles, the GA terminated, on average, after 13.19 hours only. In comparison, Gallagher~\cite{conf/cvpr/Gallagher12} reported a run time of 23.5 hours on a single 9,600-pieces puzzle, although he allows also for piece rotation. Despite achieving faster algorithms, the significant increase in run-time with respect to puzzle size has prompted us in various ways to optimize the running time before further experimenting with even larger puzzles.

One objective is to reduce considerably the running time without changing at all the GA itself. As expected, benchmark testing revealed that the most time consuming component of the GA is the creation of a new generation. This includes running the crossover procedure 1000 times on different chromosome pairs to create new offspring. Since the selection of parents and their crossover within the same generation are completely independent, creating a new generation lends itself easily to concurrent implementation, by utilizing all available CPU cores of the machine. We have implemented a parallelized version of the GA with four threads, each creating 250 offspring. Running on the 22,755- and 30,745-piece puzzle sets, this version terminated after 3.43 and 6.51 hours, respectively. In summary, our GA seems to outperform other greedy algorithms on both smaller and considerably larger puzzles.

\begin{table}[t]
\centering
\begin{tabular}{ |c||c| }
    \hline
  \# of Pieces  & Run Time \\ \hline
  \multicolumn{2}{|c|}{Single-Threaded} \\ \hline
  432 & 48.73 [sec] \\ \hline
  540 &  64.06 [sec]\\ \hline
  805 & 116.18 [sec] \\ \hline
  2,360 & 17.60 [min] \\ \hline
  3,300 & 30.24 [min] \\ \hline
  5,015 & 61.06 [min] \\ \hline
  10,375 & 3.21 [hr] \\ \hline
  22,755 & 13.19 [hr] \\ \hline
  \multicolumn{2}{|c|}{Multi-Threaded} \\ \hline
  22,755 & 3.43 [hr] \\ \hline
  30,745 & 6.51 [hr] \\ \hline

\end{tabular}
\caption{Average run times of both single-threaded and multi-threaded GA per image in every set.}
\label{tab:timesTable}
\end{table}

\subsection {Impact of each crossover phase}
Examining more carefully our 3-phase crossover operator raises the interesting question of the specific (relative) impact of each phase. To address this issue, we conducted a series of experiments, using partial crossover variants, each containing one or two of the three phases: ``agreed'', ``best-buddy'', and ``greedy''.

Certain phases may or may not assign a piece, i.e., piece assignment due to such phases is optional. For example, the ``agreed'' phase assigns a piece if it is agreed-upon by both parents; if no agreed-upon piece exists, no piece is assigned at all. Other phases, on the other hand, do guarantee piece assignment. Instead of determining whether or not to assign a piece, such phases always assign a piece from the available pool of pieces. For example, the ``greedy'' phase assigns the most compatible piece available.

Following the previous discussion, it is clear that each variant must consist of at least one assignment-guaranteed phase. Otherwise, if all optional phases do not assign a piece, the algorithm gets stuck. Since the ``greedy'' phase always assigns a piece, the implementation of all variants containing this phase is straightforward. For the other variants, we add a pseudo-phase called ``random'', which randomly assigns an available piece. Thus, unlike the ``greedy only'' variant, the ``agreed only'' variant is actually ``agreed + random''. This variant looks at all available boundaries in the growing kernel, and assigns a piece only if both parents agree on it. If no agreed-upon piece exists at any border, it randomly selects a boundary and assigns an available piece at random. (Note that this assignment creates a new boundary where an agreed-upon piece might now be placed.)

We experimented with the well-known set of Cho {\etal}~\cite{conf/cvpr/ChoAF10}, which consists of 20 432-pieces puzzles, running the above specified crossover variants of the GA on every image. The results are provided in Table~\ref{tab:phaseTable}.

\setlength{\topmargin}{22pt}

\addtolength{\textheight}{100pt}

\begin{sidewaystable}
\centering

\addvbuffer[300pt 0pt]{\begin{tabular}{|c||c|c|c||c|c|c||c|c|c|} \hline
Image \#    & agreed only & best-buddy only & greedy only & best-buddy & agreed & agreed & best-GA & avg-GA & worst-GA \\
& & & & + greedy & + greedy & + best-buddy & & &\\ \hline \hline
1   & 5.23\%        & 82.12\%           & 31.14\%       & 87.96\%     & 34.31\% & 76.03\%     & 88.44\%   & 87.55\%  & 86.62\%    \\ \hline
2   & 5.35\%        & 83.33\%           & 44.28\%       & 84.91\%     & 66.06\% & 83.09\%     & 85.28\%   & 84.73\%  & 84.18\%    \\ \hline
3   & 5.60\%        & 100.00\%          & 52.19\%       & 100.00\%    & 76.76\% & 100.00\%    & 100.00\%  & 100.00\% & 100.00\%   \\ \hline
4   & 3.53\%        & 67.03\%           & 32.97\%       & 68.49\%     & 53.41\% & 66.42\%     & 69.46\%   & 68.03\%  & 65.09\%    \\ \hline
5   & 4.26\%        & 100.00\%          & 50.24\%       & 100.00\%    & 85.89\% & 100.00\%    & 100.00\%  & 100.00\% & 100.00\%   \\ \hline
6   & 4.99\%        & 89.05\%           & 36.86\%       & 95.01\%     & 67.27\% & 90.63\%     & 98.30\%   & 97.69\%  & 96.72\%    \\ \hline
7   & 4.74\%        & 98.30\%           & 44.89\%       & 99.64\%     & 71.90\% & 99.15\%     & 100.00\%  & 100.00\% & 100.00\%   \\ \hline
8   & 4.14\%        & 100.00\%          & 57.42\%       & 100.00\%    & 89.66\% & 100.00\%    & 100.00\%  & 100.00\% & 100.00\%   \\ \hline
9   & 3.41\%        & 99.27\%           & 46.72\%       & 99.64\%     & 80.78\% & 99.15\%     & 100.00\%  & 99.82\%  & 99.64\%    \\ \hline
10  & 4.62\%        & 100.00\%          & 45.26\%       & 100.00\%    & 63.26\% & 100.00\%    & 97.81\%   & 97.81\%  & 97.81\%    \\ \hline
11  & 4.01\%        & 99.64\%           & 41.97\%       & 99.64\%     & 71.65\% & 99.64\%     & 97.08\%   & 97.08\%  & 97.08\%    \\ \hline
12  & 5.84\%        & 97.81\%           & 59.49\%       & 99.15\%     & 85.89\% & 100.00\%    & 99.64\%   & 99.39\%  & 99.03\%    \\ \hline
13  & 3.65\%        & 84.06\%           & 38.93\%       & 89.66\%     & 59.49\% & 84.43\%     & 91.12\%   & 90.56\%  & 90.15\%    \\ \hline
14  & 5.72\%        & 100.00\%          & 51.82\%       & 100.00\%    & 95.01\% & 100.00\%    & 99.64\%   & 99.64\%  & 99.64\%    \\ \hline
15  & 5.23\%        & 87.35\%           & 43.67\%       & 92.58\%     & 63.38\% & 91.61\%     & 96.84\%   & 96.35\%  & 95.86\%    \\ \hline
16  & 4.01\%        & 99.64\%           & 43.55\%       & 100.00\%    & 86.01\% & 99.15\%     & 100.00\%  & 100.00\% & 100.00\%   \\ \hline
17  & 6.20\%        & 99.64\%           & 58.03\%       & 99.64\%     & 79.81\% & 99.64\%     & 99.64\%   & 99.64\%  & 99.64\%    \\ \hline
18  & 4.87\%        & 89.90\%           & 48.30\%       & 91.85\%     & 72.75\% & 91.00\%     & 100.00\%  & 95.82\%  & 92.82\%    \\ \hline
19  & 5.11\%        & 100.00\%          & 55.60\%       & 100.00\%    & 90.51\% & 100.00\%    & 100.00\%  & 100.00\% & 100.00\%   \\ \hline
20  & 5.60\%        & 100.00\%          & 51.70\%       & 100.00\%    & 91.73\% & 100.00\%    & 100.00\%  & 100.00\% & 100.00\%   \\ \hline \hline
Avg & 4.81\%        & 93.86\%           & 46.75\%       & 95.41\%     & 74.28\% & 94.00\%     & 96.16 \%  & 95.70\%  & 95.21\%    \\ \hline
\end{tabular}}
\caption{Accuracy results for 432-piece puzzles, under the neighbor comparison, using our GA with various partial crossover variants.}
\label{tab:phaseTable}
\end{sidewaystable}

The poor results obtained by using only the ``agreed'' phase or the ``greedy'' phase are not surprising. A GA consisting solely of the ``agreed'' phase attempts to assemble the puzzle by relying, essentially, on  initial generations of random populations, {\ie} no significant crossover is expected to take place. The variant consisting only of the ``greedy'' phase is equivalent to the naive, greedy solution of the problem. On the other hand, the relatively impressive results obtained using only the ``best-buddy'' phase further underscore the effectiveness of the best-buddies concept, as presented in Pomeranz {\etal}~\cite{conf/cvpr/PomeranzSB11}.

The results of additional crossover variants, consisting only of two phases, further ascertain that the ``best-buddy'' phase seems by far the most critical. All variants containing this component score significantly higher than their counterparts. Most notably, the ``best-buddy'' phase by itself scores almost 20\% higher than the ``no best-buddy'' ({\ie} ``agreed'' + ``greedy'') variant. Again, this is not surprising, as the best-buddies criterion was shown~\cite{conf/cvpr/PomeranzSB11} to be a good indicator for the likelihood of neighboring pieces in the correct puzzle configuration. We assume that a GA based on this criterion, utilizes its many chromosomes through the generations to distinguish, eventually, between ``genuine'' and ``false'' best-buddy relations.

Having underscored the importance of the ``best-buddy'' phase, we now examine the two-phase variants containing the ``best-buddy'' component. The accuracy obtained for both of these variants is over 90\%, on average. It appears that the impact of the ``greedy'' phase is greater, as it leads to an impressive increase of 2\% when used in conjunction with the ``best-buddy'' phase. An interesting observation is that for some puzzles the GA achieves actually a better result using a partial operator variant. A more detailed examination reveals that in all of these cases, although the accuracy of the result is smaller, the fitness achieved is greater. Namely, the original operator, which uses all three phases, might find a global optimum whose score is higher than the perfect result. This is the same kind of anomaly observed with the shifted puzzles in Figure~\ref{fig:shifted}.

The final result of the GA, {\ie} the accuracy of the best chromosome in the final generation, presented in the table above, should not be regarded as a sole factor in assessing the importance of a given phase. Another important aspect is speed of convergence. Our GA always runs for 100 generations, although some puzzles could be completely solved after four or five generations. Moreover, in some puzzles, only slight changes occur in the latter generations, while in others, every generation might be as meaningful as others. Thus, for every run, we record the fitness of the best chromosome in each generation. We report the number of {\em meaningful generations}, {\ie} the number of generations for which the fitness keeps improving. We also compute the average and median improvement of the fitness through the generations. Let $F_{i}$ be the fitness of the best chromosome in the $i$-th significant generation, $\hat{F}_{i} = F_{i+1} - F_{i}$ represents the fitness improvement between the two consecutive meaningful generations $i$ and $i+1$. We report the average and median of $\hat{F}$ per image. The results appear in Table~\ref{tab:phaseGenTable}. As can be observed, the ``greedy'' phase greatly improves the speed of convergence.

\begin{sidewaystable}
\centering
\addvbuffer[300pt 0pt]{\begin{tabular}{|c||c|c|c||c|c|c||c|c|c|}
\hline
    & \multicolumn{3}{|c||}{best-buddy only}       & \multicolumn{3}{|c||}{best-buddy + greedy}   & \multicolumn{3}{|c|}{best-buddy + agreed}            \\ \hline
Image \#    & No. of significant  & Avg. fitness & Med. fitness & No. of significant  & Avg. fitness & Med. fitness   & No. of significant  & Avg. fitness & Med. fitness  \\
    & GA generations &  improvement & improvement & GA generations &  improvement & improvement &  GA generations &  improvement & improvement \\ \hline \hline
1   & 38      & 3586.15 & 1651.73 & 16        & 9089.37  & 433.65    & 32      & 3490.10 & 2639.08 \\ \hline
2   & 34      & 4642.50 & 2193.71 & 14        & 11751.01 & 13.00     & 46      & 3392.00 & 1582.93 \\ \hline
3   & 29      & 4775.30 & 2710.44 & 6         & 27306.71 & 10208.91  & 29      & 4906.21 & 2715.83 \\ \hline
4   & 36      & 3885.35 & 1106.94 & 11        & 14914.72 & 2423.29   & 25      & 5569.29 & 3664.47 \\ \hline
5   & 28      & 5952.24 & 4742.76 & 6         & 32208.79 & 6950.50   & 32      & 5244.04 & 2059.36 \\ \hline
6   & 38      & 3488.45 & 3089.90 & 16        & 9067.55  & 421.01    & 46      & 2912.19 & 1763.02 \\ \hline
7   & 53      & 3479.68 & 1826.83 & 9         & 22375.16 & 1773.32   & 44      & 4190.63 & 2705.12 \\ \hline
8   & 31      & 3439.34 & 1367.11 & 5         & 25793.19 & 17396.02  & 33      & 3184.64 & 2186.03 \\ \hline
9   & 38      & 3802.34 & 2060.44 & 11        & 14152.20 & 72.10     & 39      & 3634.47 & 1185.56 \\ \hline
10  & 38      & 2777.28 & 1845.33 & 6         & 20229.33 & 10828.36  & 39      & 2619.01 & 1958.74 \\ \hline
11  & 28      & 2341.76 & 1868.33 & 6         & 12747.09 & 5100.99   & 33      & 1937.43 & 1846.94 \\ \hline
12  & 36      & 3699.86 & 1840.87 & 11        & 12989.78 & 62.29     & 39      & 3370.68 & 917.29  \\ \hline
13  & 42      & 2834.61 & 1406.46 & 11        & 12092.63 & 1372.75   & 34      & 3516.87 & 1966.51 \\ \hline
14  & 33      & 3436.79 & 2577.28 & 5         & 27851.13 & 14883.22  & 29      & 4012.27 & 1648.06 \\ \hline
15  & 45      & 3599.74 & 2032.63 & 10        & 18018.68 & 2121.83   & 37      & 4437.38 & 2911.77 \\ \hline
16  & 34      & 3995.88 & 3334.25 & 6         & 26835.21 & 10911.81  & 34      & 4039.69 & 3742.05 \\ \hline
17  & 33      & 4325.24 & 1845.80 & 8         & 19592.13 & 1252.53   & 25      & 5850.35 & 3285.79 \\ \hline
18  & 35      & 2371.56 & 1507.59 & 10        & 9083.49  & 514.49    & 40      & 2103.28 & 1260.12 \\ \hline
19  & 22      & 6590.41 & 4616.78 & 5         & 33977.85 & 18104.56  & 23      & 6211.21 & 4461.81 \\ \hline
20  & 26      & 4200.16 & 1441.01 & 6         & 20946.65 & 4308.45   & 28      & 3868.74 & 2784.35 \\ \hline \hline
Avg & 34.85   & 3861.23 & 2253.31 & 8.90      & 19051.13 & 5457.65   & 34.35   & 3924.52 & 2364.24 \\ \hline
\end{tabular}}
\caption{Convergence speed of the GA with different partial crossover variants. For each image we report the number of meaningful generations and the average and median of fitness improvement through the generations. }
\label{tab:phaseGenTable}
\end{sidewaystable}

In conclusion, it appears that all phases play a significant role, assisting both the accuracy achieved and the speed of convergence. In most cases, the original operator scores higher and even manages to obtain better-than-perfect scores. These results give rise to many other interesting questions such as the impact of the phases on harder puzzles or the elusive contribution of the agreed phase. We leave these questions for future research.

\subsection{Different fitness function}

It is well known that a GA relies greatly on the quality of its fitness measure; in the context of the jigsaw puzzle problem, a good fitness measure should be correlative to the piece configuration and allow also for convergence. The previously described anomaly of the shifted puzzles calls for the examination of different fitness functions, as the one used by us proved inexact in some cases. Recent research by Gallagher~\cite{conf/cvpr/Gallagher12} suggests the {\em Mahalanobis gradient compatibility} (MGC) as a preferable compatibility measure to those used by Pomeranz {\etal}~\cite{conf/cvpr/PomeranzSB11}. The MGC penalizes changes in intensity gradients, rather than changes in intensity, and learns the covariance of the color channels, using the Mahalanobis distance. Critics of dissimilarity suggest that the introduction of these properties in the MGC measure could help alleviate the shifting phenomenon.

We ran the GA on the 20-image set (of 432-piece puzzles) supplied by Cho {\etal}~\cite{conf/cvpr/ChoAF10}, replacing the dissimilarity fitness function with the above MGC measure. Incidentally, this particular setting enables also a more meaningful comparison with the performance of Gallagher's algorithm (95.1\% accuracy on the above image set), which uses, of course, the MGC measure. Experimental results are given in Table~\ref{tab:galTable}. Employing the GA with the MGC measure improves the average accuracy achieved for this set from 95.1\% (achieved by Gallagher) to 96.03\%. Again, this attests to the superior performance of the GA-based solver.

\begin{table}[t]
\centering
\begin{tabular}{|c||c|c|} \hline
Image \# & GA-MGC    & GA-Dissimiliarity \\ \hline \hline
1   & 87.10  & 88.44 \\ \hline
2   & 82.97  & 85.28 \\ \hline
3   & 100.00 & 100.00 \\ \hline
4   & 65.69  & 69.46 \\ \hline
5   & 100.00 & 100.00 \\ \hline
6   & 100.00 & 98.30  \\ \hline
7   & 100.00 & 100.00  \\ \hline
8   & 100.00 & 100.00 \\ \hline
9   & 100.00 & 100.00 \\ \hline
10  & 100.00 & 97.81 \\ \hline
11  & 99.64  & 97.08  \\ \hline
12  & 99.27  & 99.64 \\ \hline
13  & 91.85  & 91.12 \\ \hline
14  & 100.00 & 99.64 \\ \hline
15  & 94.04  & 96.84 \\ \hline
16  & 100.00 & 100.00 \\ \hline
17  & 100.00 & 99.64  \\ \hline
18  & 100.00 & 100.00  \\ \hline
19  & 100.00 & 100.00 \\ \hline
20  & 100.00 & 100.00 \\ \hline \hline
Avg & 96.03  & 96.16  \\ \hline
\end{tabular}
\caption{Accuracy results using our GA with MGC fitness measure vs. dissimilarity fitness.}
\label{tab:galTable}
\end{table}

\begin{figure*}
\centering
        \begin{subfigure}[t]{0.30\textwidth}
                \centering
                \includegraphics[width=\textwidth]{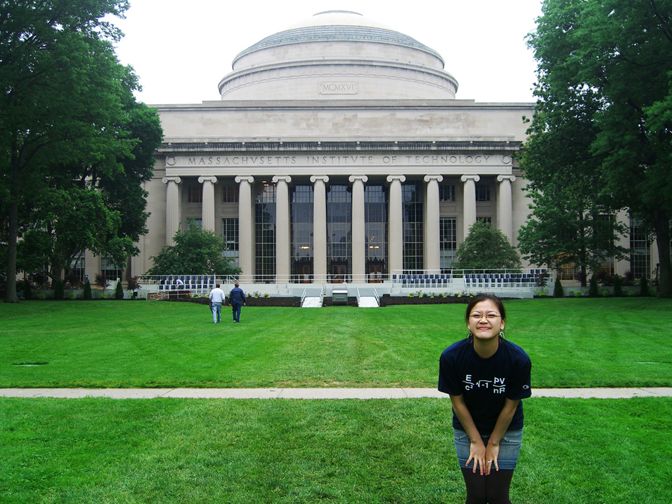}
                \caption{Original image}
                \label{fig:result_10375_04_gen_00000000}
        \end{subfigure}%
        ~ 
        \begin{subfigure}[t]{0.30\textwidth}
                \centering
                \includegraphics[width=\textwidth]{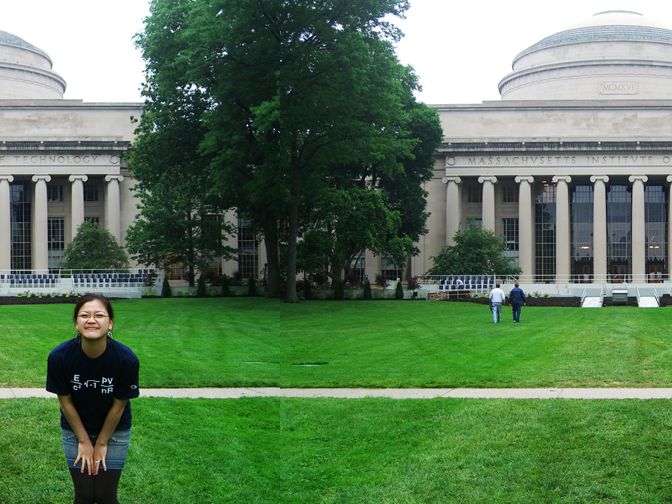}
                \caption{Generation 6}
                \label{fig:result_10375_04_gen_00000001}
        \end{subfigure}
        ~ 
        \begin{subfigure}[t]{0.30\textwidth}
                \centering
                \includegraphics[width=\textwidth]{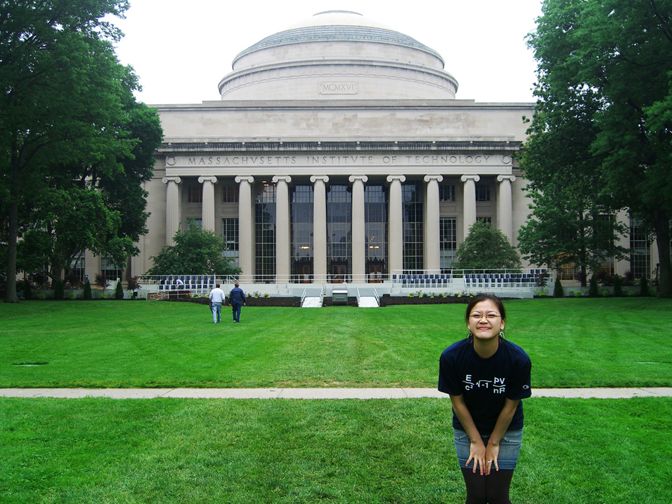}
                \caption{Generation 9}
                \label{fig:result_10375_04_gen_00000002}
        \end{subfigure}
        ~
        \begin{subfigure}[t]{0.30\textwidth}
                \centering
                \includegraphics[width=\textwidth]{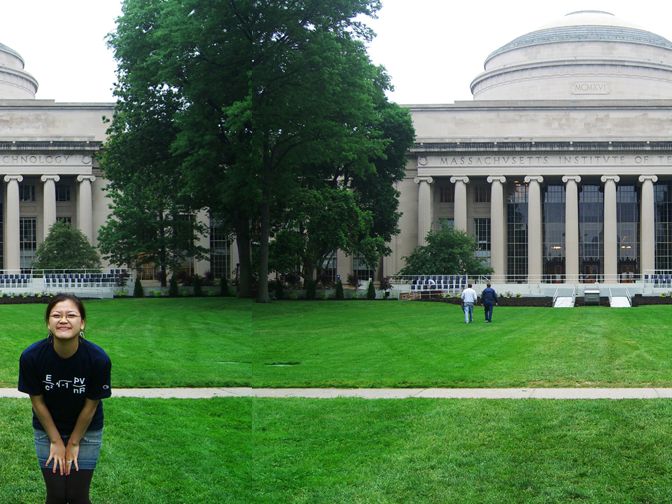}
                \caption{Generation 10}
                \label{fig:result_10375_04_gen_00000100}
        \end{subfigure}
        ~
        \begin{subfigure}[t]{0.30\textwidth}
                \centering
                \includegraphics[width=\textwidth]{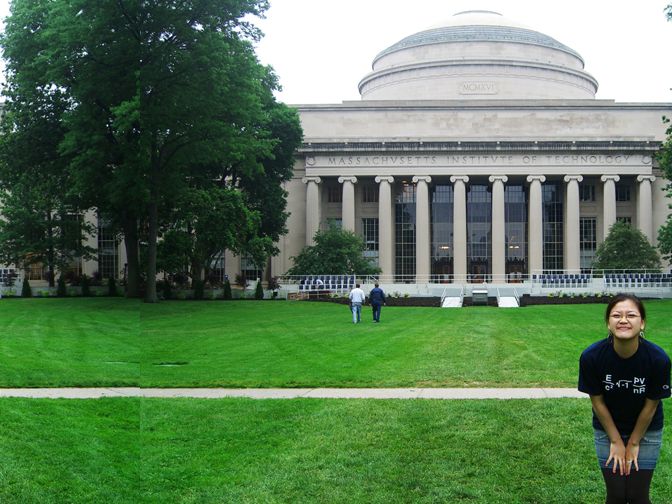}
                \caption{Generation 16 (final)}
                \label{fig:result_10375_04_gen_00000100}
        \end{subfigure}
        \caption{Shifted puzzle solution obtained using the GA with MGC fitness function; best chromosome from selected generations is shown, each with a higher fitness than preceding chromosomes.}
        \label{fig:mgcShifted}
\end{figure*}

Unlike other greedy solvers, which use similar compatibility measures, our GA might yield, on occasion, a ``better-than-perfect'' result, {\ie} an improved fitness value for an incorrect piece configuration. This anomaly could occur, for example, by rearrangement of unicolor surfaces (like the skies) or a misplacement of two or more correctly assembled segments (as illustrated in Figure~\ref{fig:shifted}), due to a high color similarity between their abutting edges ({\eg} a sea positioned above the skies). Despite the drawback of the fitness function used, the above phenomenon actually attests to the power of the GA, as it optimizes the fitness function to an extent that has not been observed or reported before. To further explore this phenomenon, we compare between the original dissimilarity-based GA and a GA variant based on the MGC measure. Shifted puzzle solutions, using the original GA version, are now solved correctly. On the other hand, a few images solved correctly with the dissimilarity fitness, are now shifted, using the MGC measure. Figure~\ref{fig:mgcShifted} illustrates shifted solutions obtained in various generations using the GA with the MGC measure.

\section{Discussion and future work}

In this paper we presented a highly effective jigsaw puzzle solver, based for the first time on evolutionary computation. The automatic solver developed is capable of reconstructing puzzles of up to 30,745 pieces ({\ie} more than twice the puzzle size that has been attempted before). The performance observed is comparable, if not superior to other known solvers, in most of the cases examined. Also, we created new sets of very large images for benchmark testing of this and other solvers, and supplied both the image sets and our results for the benefit of the community~\cite{conf/cvpr/site/Our}.

By introducing a novel crossover technique, we were able to arrive at a most effective solver. Our approach could prove useful in further utilization of GAs ({\eg},~\cite{sholomon2014generalized,sholomon2014genetic}) for solving more difficult variations of the jigsaw problem (including unknown piece orientation, missing and excessive puzzle pieces, unknown puzzle dimensions, and three-dimensional puzzles), and could also assist in the design of GAs in other problem domains.


\bibliographystyle{spmpsci}      
\bibliography{egbib}   

\end{document}